\documentclass[10pt,twocolumn,letterpaper]{article}

\usepackage{iccv}
\usepackage{times}
\usepackage{epsfig}
\usepackage{graphicx}
\usepackage{amsmath}
\usepackage{amssymb}
\usepackage{multirow}
\usepackage{algorithm}
\usepackage{algorithmic}
\usepackage[table,xcdraw]{xcolor}

\usepackage[pagebackref=true,breaklinks=true,letterpaper=true,colorlinks,bookmarks=false]{hyperref}

\iccvfinalcopy 

\def\iccvPaperID{4472} 

\ificcvfinal\fi

\begin{document}

\title{Robust GAN inversion}

\author{Egor Sevriugov\\
Skolkovo institute of science and technology\\
Moscow, Russia 121205\\
{\tt\small egor.sevriugov@skoltech.ru}
\and
Ivan Oseledets\\
Skolkovo institute of science and technology\\
Moscow, Russia 121205\\
{\tt\small I.Oseledets@skoltech.ru}
}

\maketitle
\ificcvfinal\thispagestyle{empty}\fi

\begin{abstract}
Recent advancements in real image editing have been attributed to the exploration of Generative Adversarial Networks (GANs) latent space. However, the main challenge of this procedure is GAN inversion, which aims to map the image to the latent space accurately. Existing methods that work on extended latent space $W+$ are unable to achieve low distortion and high editability simultaneously. To address this issue, we propose an approach which works in native latent space $W$ and tunes the generator network to restore missing image details. We introduce a novel regularization strategy with learnable coefficients obtained by training randomized StyleGAN 2 model - WRanGAN. This method outperforms traditional approaches in terms of reconstruction quality and computational efficiency, achieving the lowest distortion with 4 times fewer parameters. Furthermore, we observe a slight improvement in the quality of constructing hyperplanes corresponding to binary image attributes. We demonstrate the effectiveness of our approach on two complex datasets: Flickr-Faces-HQ and LSUN Church.
\end{abstract}

\section{Introduction}
The emergence of generative adversarial neural networks (GANs) has made a great contribution to high quality image synthesis. A well-known model in this field is StyleGAN, which has achieved remarkable results. Moreover, several works \cite{pfau2020disentangling,khrulkov2021disentangled,peebles2020hessian,voynov2020unsupervised,shen2020interpreting,disentangledrepr} have demonstrated that GANs possess a wide range of interpretable semantics, providing the basis for image editing. This property enables the alteration of certain attributes while preserving the identity of the image relative to others. However, the application of this property to real images has been limited due to the need to accurately map them into the latent space. This task, known as GAN inversion, initially focused on mapping images into the native latent space $W$. Yet, authors in \cite{abdal2019image2stylegan} have shown that this approach leads to significant differences between the original and generated images. Subsequent work has shifted focus to the extended latent space $W+$\cite{Abdal_2020_CVPR,Richardson_2021_CVPR,alaluf2021restyle,Tov2021Designing,wang2021HFGI,hu2022style}, which improves the quality of image reconstruction but degrades editability. This issue, called the distortion-editability tradeoff \cite{Tov2021Designing}, limits the possibility of using codes obtained in the $W+$ space.

Another way to solve the problem was proposed in \cite{roich2021pivotal}, which includes a small change in the generator parameters when working with the latent space $W$ - pivotal tuning inversion (\textbf{PTI}). We improve their idea by using adaptive regularization instead of one number, 
since each parameter has different contribution to the model performance.

In this paper, we present a novel approach for learning regularization that allows for high-quality image reconstructions while preserving the ability of the model to generate realistic images. This approach is based on a randomized version of StyleGAN 2 called WRanGAN, in which part of the model weights are assumed to be normally distributed with trainable mean and variance. To apply different regularization coefficients, we use the reparameterization trick \cite{Kingma2014AutoEncodingVB} during the inversion procedure. The effectiveness of our technique was evaluated on two complex datasets, the Flickr-Faces-HQ Dataset (FFHQ) \cite{karras2019stylebased} and LSUN Churches \cite{yu15lsun}. Our contributions are summarized as follows:
\begin{itemize}
\item We present a novel adaptive regularization scheme based on an investigation of different regularization strategies and their effect on reconstruction quality and model corruption.
\item We introduce \textbf{WRanGAN}, a model that learns appropriate regularization coefficients via a randomization of the StyleGAN model.
\item We evaluate \textbf{WRanGAN} in terms of generation, reconstruction, binary attributes extraction and computational cost, and compare it to several baselines in a qualitative and quantitative manner.
\end{itemize}

\section{Problem setting}
\subsection{Latent Space Manipulation}


GANs allow the generation of images that are controlled by semantic directions \cite{pfau2020disentangling,khrulkov2021disentangled,peebles2020hessian,voynov2020unsupervised,shen2020interpreting,disentangledrepr}. In particular, in the work \cite{disentangledrepr} authors proposed estimating the subspaces that are invariant under random-walk diffusion for identification. Supervision in the form of facial attribute labels was used in \cite{shen2020interpreting} to find meaningful linear directions in the latent space. The identification of latent directions based on the principal component analysis (PCA) was proposed in \cite{disentangledrepr}.

\subsection{GAN inversion}

Recent research has focused on the improving reconstruction quality of GAN inversion task, which involves finding the latent code that accurately reproduce real image. This task can be divided into two main groups: optimization methods that directly modify the latent code to minimize a loss function \cite{abdal2019image2stylegan,maskedgan}, and encoder-based methods that use a trained encoder to generate an image \cite{guan2020collaborative,alaluf2021restyle,Richardson_2021_CVPR,Tov2021Designing}. Generally, methods operate in the native latent space $W$, which can lead to significant visual differences compared to the original image \cite{abdal2019image2stylegan}. On the other hand, the extended latent space $W+$ is much more expressive and allows for the reproduction of more unique image details. However, this approach is limited by the fixed generator parameters. To address this issue, some approaches have proposed to modify the generator network to fix visual artifacts, as demonstrated in \cite{roich2021pivotal}. Others have used hypernetworks to predict the change of generator parameters in order to minimize distortion and preserve the realism of the generated image, as seen in \cite{alaluf2021hyperstyle} and \cite{dinh2021hyperinverter}.

\subsection{Distortion-editability tradeoff}


 The GAN inversion in the extended latent space $W+$ significantly improves reconstruction of real images, but at the same time it leads to degradation of editability called distortion-editability trade-off \cite{Tov2021Designing}. There are works \cite{zhu2020indomain} and \cite{Tov2021Designing}, where the authors proposed to search for editable latent codes in an extended latent space $W+$. A completely different way to solve this problem was proposed in the works \cite{alaluf2021hyperstyle} and \cite{roich2021pivotal}. Instead of trying to find a balance between editability and distortion, the authors suggest using the advantage of projection into the latent space $W$ and updating the generator parameters to minimize distortion. In this paper we also used projection to native latent space to reach high editability.
\begin{figure*}[t]
   \centering
    \includegraphics[width=0.95\linewidth]{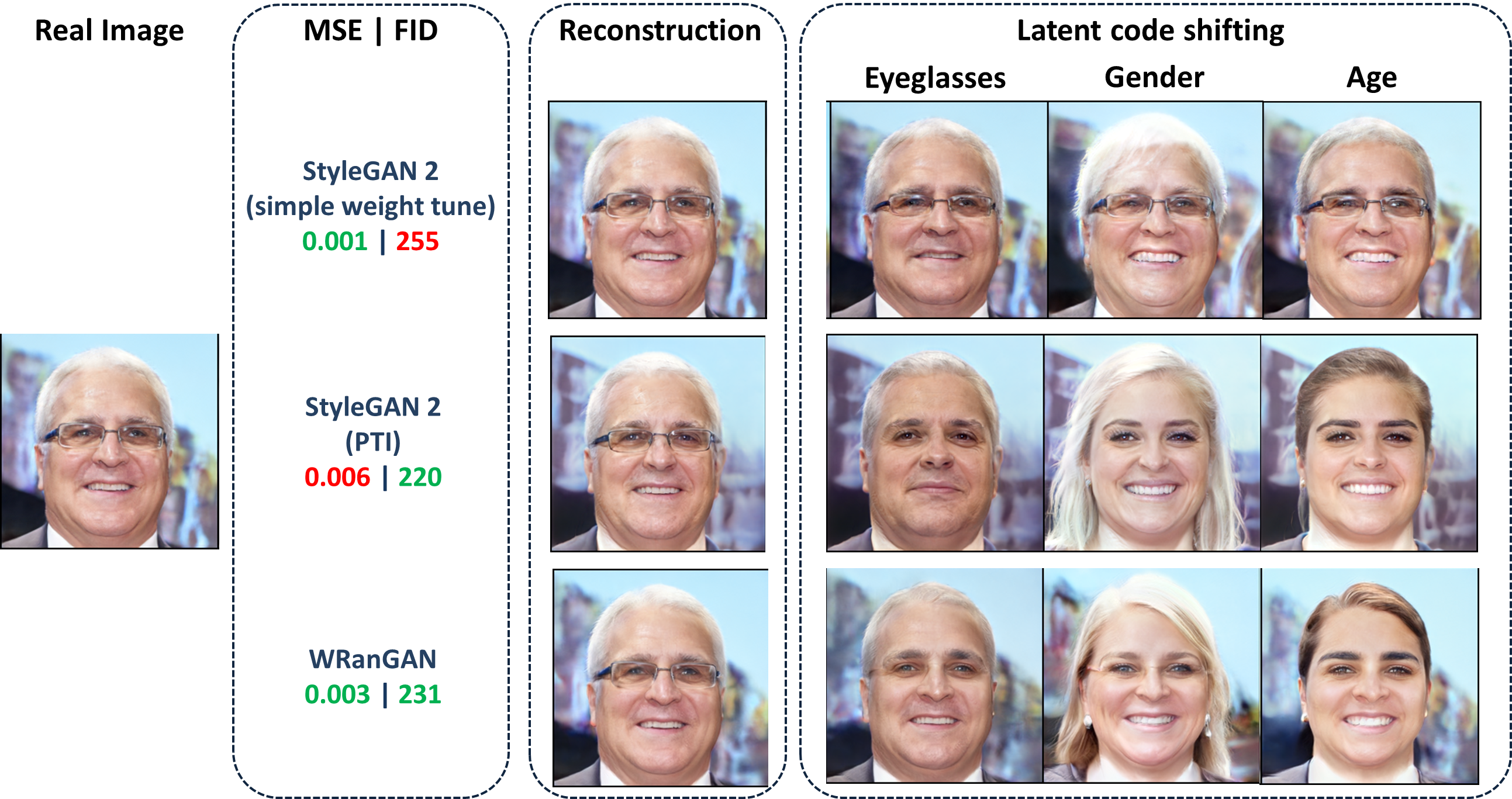}
    \caption{Comparison of different regularization strategies of GAN inversion. \textbf{Simple Weight Tune} represents optimization of model parameters with low regularization coefficient. \textbf{PTI} represents pivotal tuning approach - high regularization coefficient. And the last one is proposed \textbf{\textbf{WRanGAN}}. For each approach metrics were calculated: MSE (lower values is better) measuring distortion, FID (lower values is better) evaluated over images generated by shifting latent code, measuring model corruption. For each approach we presented result of latent code shifting in the 4 orthogonal directions corresponding to the maximal change in the image.}
    \label{fig:rec_real_tradeoff}
\end{figure*}
\subsection{Generator tuning}

Model tuning significantly improves ability to reproduce real image \cite{roich2021pivotal,alaluf2021hyperstyle,dinh2021hyperinverter}. But changing the parameters of the model can damage its quality. In order to improve the realism of generated images after modification of the generator weights, non-saturating GAN loss was used to train hypernetworks \cite{alaluf2021hyperstyle,dinh2021hyperinverter} (encoders predicting the necessary weight shift) . Despite the significant improvement in the quality of reproduction, these methods are still inferior to the PTI approach \cite{roich2021pivotal} based on direct weight optimization. But optimization of model parameters without any additional constraints requires imposing a regularization with a high coefficient in order not to damage the realism of the generated images and forces to optimize all the parameters of the model to reach low distortion. As a result, it leads to a significant increase in the computational costs.



\section{Method}

In general, the proposed method solves the general problem by applying non-equal learnable regularization. This allows to set appropriate regularization coefficient for each parameter depending on its effect on model performance (realism of generated images). In general approach, the first stage learns appropriate regularization coefficients for the inversion task by adversarial training of a generator with partially randomized parameters. The second stage, then, uses these coefficients for an inversion procedure consisting of encoder projection and regularized optimization minimizing particular loss function.

\begin{figure}[h!]
   \centering
    \includegraphics[width=1.0\linewidth]{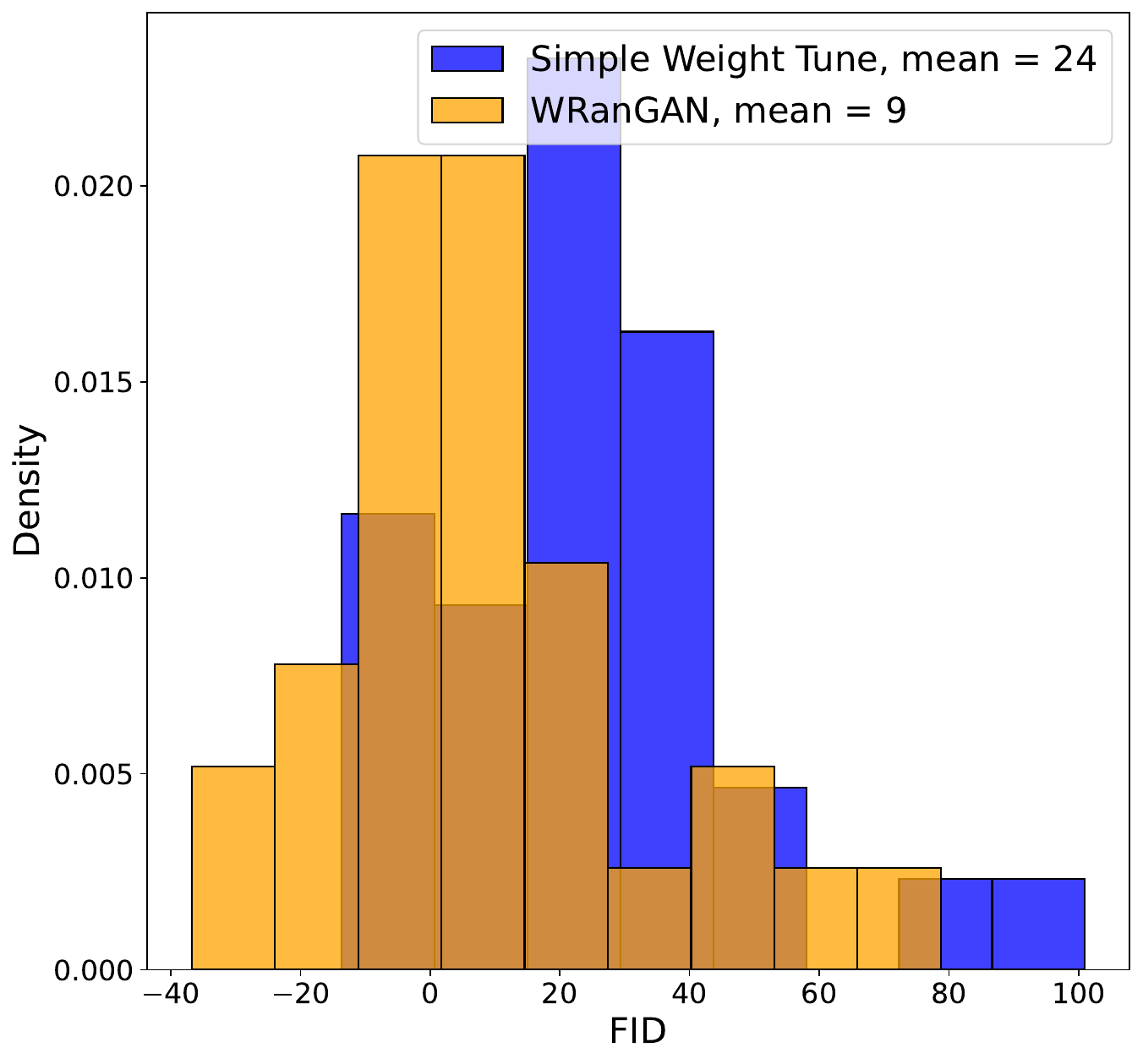}
    \caption{Model corruption evaluation for two regularization strategies: \textbf{Simple Weight Tune} and \textbf{WRanGAN}. Calculations performed over 30 randomly taken images from FFHQ dataset. Corresponding mean values are also presented.}
    \label{fig:fid_comp}
\end{figure}

\subsection{Model corruption assessment}


To measure how well the model after tuning is, we use a common technique known as Fréchet Inception Distance (FID) \cite{heusel2018gans} and Kernel Inception Distance (KID) \cite{binkowski2018demystifying}. In general case, evaluation is performed on a large set of images produced by the generator network. However, since our main focus is on edited images, we slightly change this tool by performing calculations over 1000 images obtained by shifting latent code (editing) in random directions. The shift norm is taken accordingly to the characteristic size of the style space (variance of style features).

\subsection{Regularized inversion}

In order to avoid degradation of realism in generated images, regularization term is often added to optimization procedure. The general task formulation looks like this:

$$\hat{w},\hat{\theta}_G = \arg\min\limits_{w,\theta_G} \mathcal{L}(G(w,\theta_G),\hat{x}) + \alpha_{\mathrm{reg}}\|\theta_G-\theta_{G,0}\|^2_2$$


Here, $\hat{x}$ some real image, $\alpha_{\mathrm{reg}}\|\theta_G-\theta_{G,0}\|^2_2$ the regularization term, $\theta_{G,0}$ the initial values of the generator weights. For \textbf{WRanGAN} inversion we used $\mathcal{L} = 2\mathcal{L}_2 + \mathcal{L}_{\mathrm{LPIPS}}$ and initialize intermediate latent code $w$ by mapping the output of the trained encoder $E$ to intermediate latent space $W$: $w = f(E(\hat{x}))$.  $\alpha_{\mathrm{reg}}$ - regularization coefficient, which choice is the balance between reconstruction quality and model corruption. We considered three strategies of regularization to illustrate this paradigm:
\begin{itemize}
    \item low regularization value (\textbf{Simple Weight Tune})
    \item high regularization value (\textbf{PTI})
    \item appropriate regularization coefficients (\textbf{WRanGAN})
\end{itemize}



The results of our experiments, presented in Figure~\ref{fig:rec_real_tradeoff}, demonstrate that a low regularization value can impair visual quality and significantly reduce the variability of the model (the ability to manipulate certain image attributes suffers significantly). This is reflected in the FID metric used for model corruption evaluation, which shows the highest value for \textbf{Simple Weight Tune} strategy. Furthermore, applying a high regularization coefficient does not reach the lowest distortion, as evidenced by the mean squared error (MSE). Finally, the proposed \textbf{WRanGAN} model shows a good balance between both aspects: reconstruction quality and model corruption.

A more detailed statistical view is presented in Figure~\ref{fig:fid_comp}. Here, we evaluated the model corruption for the \textbf{Simple Weight Tune} and \textbf{WRanGAN} strategies in comparison with the \textbf{PTI} approach - the difference for each image between the FID metric value for the particular strategy and the corresponding value for \textbf{PTI}. The results for the appropriate regularization strategy are much better than those for \textbf{Simple Weight Tune}. The proposed method does not corrupt the model.

\subsection{\textbf{WRanGAN} inversion}



In this part we have discussed how to apply regularization to randomized model parameters $\theta_G \sim N(\mu_\theta,\sigma_\theta)$. To this end, we employ the reparameterization trick, which states that $\theta_G^i = \mu_\theta^i + \epsilon^i \sigma_\theta^i$ where $\epsilon \sim N(0,1)$ and $i$ is the index of a particular parameter. By regularizing the parameter $\epsilon$, we get the following equation: $$\alpha_{\mathrm{reg}}\|\epsilon\|^2_2 = \sum\limits_{i}\frac{\alpha_{\mathrm{reg}}}{\sigma^i_\theta}(\theta^i_G - \mu^i_\theta)^2 = \sum\limits_i\alpha_{\mathrm{reg}}^i(\theta_G^i - \theta^i_{G,0})^2$$ Here, we have used the notations $\alpha_{\mathrm{reg}}^i = \frac{\alpha_{\mathrm{reg}}}{\sigma^i_\theta}$ and $\mu^i_\theta = \theta^i_{G,0}$, and have obtained a standard regularization formulation with different regularization coefficients for each randomized model parameter. The tips discussed in this part are summarized in Algorithm~\ref{alg:GAN_inversion}.

\begin{algorithm}[t]
\caption{Algorithm of WRanGAN inversion}
\textbf{Input}: real image $\hat{x}$, generator parameters $\mu_\theta,\sigma_\theta$\\
\textbf{Parameter}: regularization coefficient $\alpha_{\mathrm{reg}}$\\
\textbf{Output}: latent code $w$ and parameterized randomization $\epsilon$\\
\begin{algorithmic}

\STATE Initialize $w = E(x)$ by the output of encoder network
\STATE Initialize $\epsilon$ with small value ($10^{-4}$)
\FOR{number of iterations}
\STATE Set generator weights $\theta_G \leftarrow \mu_\theta + \sigma_\theta \epsilon$
\STATE Update parameters $w, \epsilon$ minimizing:
\[\mathcal{L}(G(w, \theta_G),\hat{x}) + \alpha_{\mathrm{reg}}\|\epsilon\|_2^2\]

\ENDFOR
\STATE \textbf{return} $w$,$\epsilon$
\end{algorithmic}
\label{alg:GAN_inversion}
\end{algorithm}

\begin{figure}[t]
   \centering
    \includegraphics[width=1.0\linewidth]{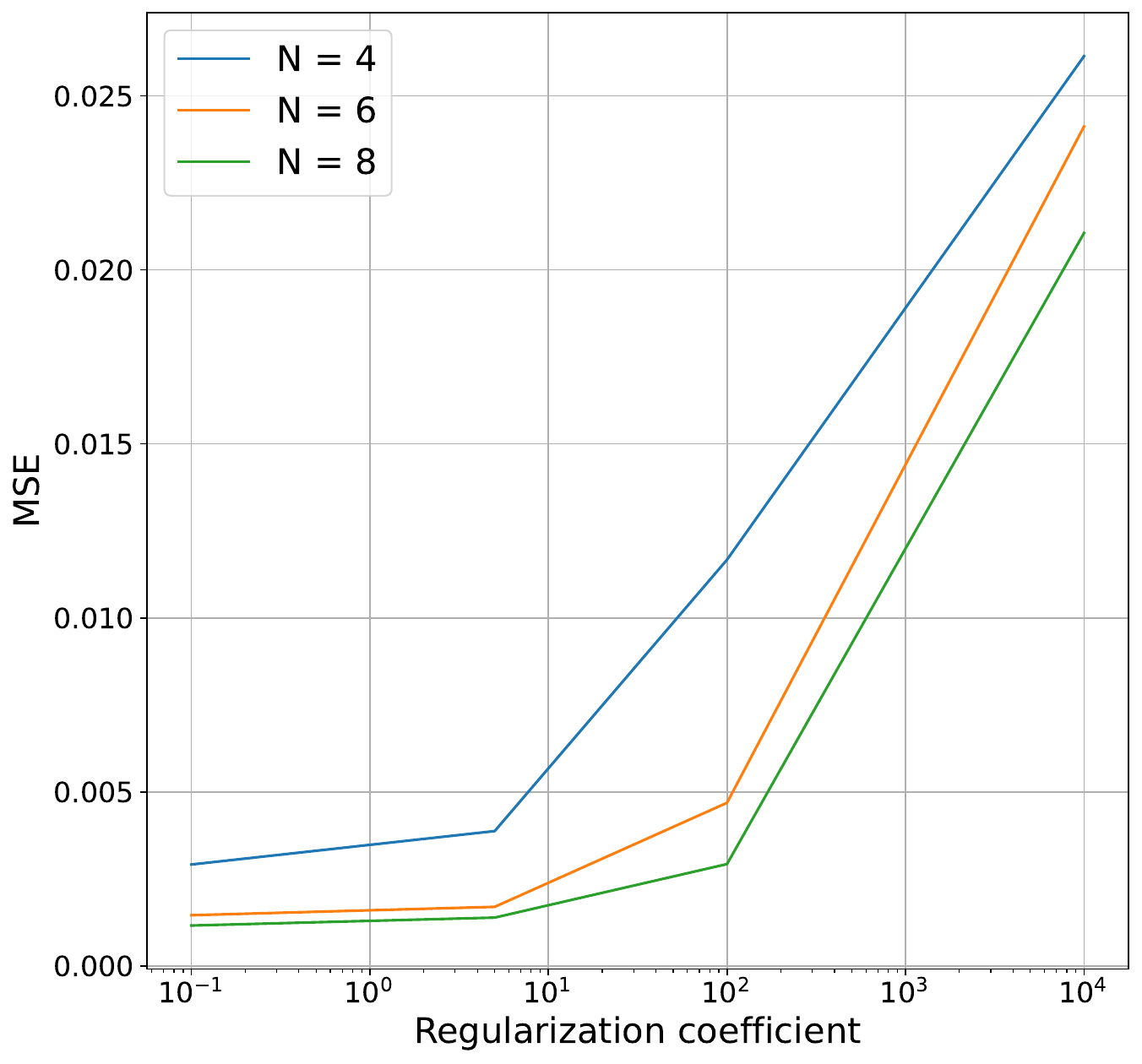}
    \caption{Dependence of MSE on number of randomized layers $N$ versus regularization coefficient. The lower the curve the better chosen number of layers.}
    \label{fig:inv_layer_reg}
\end{figure}


\begin{algorithm}[t]
\caption{WRanGAN training algorithm.}
\textbf{Input}: pretrained StyleGAN 2 weights $\theta_{G,0}$, dataset $\hat{x}$\\
\textbf{Parameter}: batch size $m$\\
\textbf{Output}: $\mu_\theta$, $\sigma_\theta$\\
\begin{algorithmic}
\STATE Initialize $\mu_\theta = \theta_{G,0}$
\STATE Initialize $\sigma_\theta = 1$ for randomized parameters
\FOR{number of training iterations}
\STATE Sample ${z^{(1)}, ..., z^{(m)}} \sim N(0,1)$
\STATE Map to intermediate latent space $w^{(i)} = f(z^{(i)})$
\STATE Sample ${\hat{x}^{(1)}, ..., \hat{x}^{(m)}}$ from training dataset $\hat{x}$
\STATE Sample $\epsilon \sim N(0,1)$ and calculate $\theta_G = \mu_\theta + \epsilon \sigma_\theta$
\STATE Update discriminator weights $\theta_D$ minimizing:
\[\frac{1}{m}\sum\limits_{i=1}^m\mathcal{L}_D(D(\hat{x}^{(i)}), D(G(w^{(i)}, \theta_G)))\]

\STATE Sample ${z^{(1)}, ..., z^{(m)}} \sim N(0,1)$
\STATE Map to intermediate latent space $w^{(i)} = f(z^{(i)})$
\STATE Sample $\epsilon \sim N(0,1)$ and calculate $\theta_G = \mu_\theta + \epsilon \sigma_\theta$
\STATE Update parameters $(\mu_\theta, \sigma_\theta)$ minimizing:
\[\frac{1}{m}\sum\limits_{i=1}^m\mathcal{L}_g(D(G(w^{(i)}, \theta_G)))\]

\ENDFOR
\STATE \textbf{return} $\mu_\theta$,$\sigma_\theta$
\end{algorithmic}
\label{alg:GAN_training}
\end{algorithm}

\begin{table}[t]
  \centering
  \begin{tabular}{|c|c|}
\hline
\textbf{\begin{tabular}[c]{@{}c@{}}Number of\\randomized layers\end{tabular}} &  \textbf{\begin{tabular}[c]{@{}c@{}}Relative increase in\\ amount of parameters\end{tabular}}                                                                 \\ \hline
4                                                                                                                                                                 & 7\%                                                                              \\ \hline
6                                                                                                                                                                & 23\%                                                                             \\ \hline
8                                                                                                                                                                 & 39\%                                                                             \\ \hline
\end{tabular}
\caption{Quantitative comparison of memory cost on the number of randomized layers}
\label{tab:comp_n_rand_layers}
\end{table}

\begin{table*}[t]
\centering
\begin{tabular}{c|c|c|cccccc}
\hline
                                                                        &                                  &                                                             &                                & \multicolumn{2}{c}{\textbf{LPIPS}$\downarrow$} &                                    &                                                                                     &                                                                               \\
\multirow{-2}{*}{\textbf{Domain}}                                       & \multirow{-2}{*}{\textbf{Model}} & \multirow{-2}{*}{\textbf{Method}}                           & \multirow{-2}{*}{\textbf{MSE}$\downarrow$} & \textbf{VGG}     & \textbf{Alex}   & \multirow{-2}{*}{\textbf{MS-SSIM}$\uparrow$} & \multirow{-2}{*}{\textbf{\begin{tabular}[c]{@{}c@{}}GPU usage\\ (MB)$\downarrow$\end{tabular}}} & \multirow{-2}{*}{\textbf{\begin{tabular}[c]{@{}c@{}}Time\\ (s)$\downarrow$\end{tabular}}} \\ \hline
                                                                        &                                  & E4E                                                         & 0.062                          & 0.389            & 0.235           & 0.605                              & 2499                                                                                & 1.64                                                                          \\
                                                                        &                                  & Restyle                                                     & 0.035                          & 0.335            & 0.154           & 0.72                               & \textbf{2483}                                                                       & \textbf{0.28}                                                                 \\
                                                                        &                                  & SG2 W+                                                      & 0.04                           & 0.14             & 0.138           & 0.783                              & 4295                                                                                & 97.9                                                                          \\
                                                                        &                                  & HyperStyle                                                  & 0.026                          & 0.288            & 0.105           & 0.788                              & 3583                                                                                & 0.31                                                                          \\
                                                                        & \multirow{-5}{*}{StyleGAN 2}     & PTI                                                         & 0.024                          & 0.293            & \textbf{0.06}   & 0.776                              & 3133                                                                                & 35.46                                                                         \\ \cline{2-9} 
\multirow{-6}{*}{FFHQ}                                                  & WRanGAN                          & \begin{tabular}[c]{@{}c@{}}WRanGAN\\ inversion\end{tabular} & \textbf{0.007}                 & \textbf{0.085}   & 0.083           & \textbf{0.929}                     & {\color[HTML]{3531FF} 2557}                                                         & {\color[HTML]{3531FF} 23.27}                                                  \\ \hline
                                                                        &                                  & E4E                                                         & 0.142                          & 0.506            & 0.418           & 0.263                              & 2499                                                                                & 1.64                                                                          \\
                                                                        &                                  & Restyle                                                     & 0.087                          & 0.411            & 0.25            & 0.489                              & \textbf{2483}                                                                       & \textbf{0.28}                                                                 \\
                                                                        &                                  & SG2 W+                                                      & 0.107                          & 0.225            & 0.235           & 0.543                              & 4295                                                                                & 97.9                                                                          \\
                                                                        & \multirow{-4}{*}{StyleGAN 2}     & PTI                                                         & 0.053                          & 0.411            & \textbf{0.065}  & 0.643                              & 3133                                                                                & 47                                                                            \\ \cline{2-9} 
\multirow{-5}{*}{\begin{tabular}[c]{@{}c@{}}LSUN\\ Church\end{tabular}} & WRanGAN                          & \begin{tabular}[c]{@{}c@{}}WRanGAN\\ inversion\end{tabular} & \textbf{0.033}                 & \textbf{0.177}   & 0.224           & \textbf{0.782}                     & {\color[HTML]{3531FF} 2557}                                                         & {\color[HTML]{3531FF} 23.27}                                                  \\ \hline
\end{tabular}
\caption{Quantitative reconstruction results of \textbf{WRanGAN} model with corresponding method compared to StyleGAN 2 inversion approaches including encoder and optimization based. Assessment performed over several standard metrics, for each of them, the arrow identifies which values are better (lower $\downarrow$ / higher $\uparrow$). The best results for each evaluated metric are highlighted in \textbf{bold}. Values in \textcolor{blue}{blue} outline the cases that we outperform PTI.}
\label{tab:rec_comp}
\end{table*}

\begin{table}[t]
\centering
\begin{tabular}{c|c|ccc}
\hline
\textbf{Domain}                                                        & \textbf{Model} & \textbf{FID}  & \textbf{Precision} & \textbf{Recall} \\ \hline
\multirow{2}{*}{\begin{tabular}[c]{@{}c@{}}Human\\ Faces\end{tabular}} & StyleGAN 2     & \textbf{4.27} & \textbf{0.7}       & 0.42            \\
                                                                       & WRanGAN        & 5.61          & 0.65               & \textbf{0.45}   \\ \hline
\multirow{2}{*}{\begin{tabular}[c]{@{}c@{}}LSUN\\ Church\end{tabular}} & StyleGAN 2     & 4.3           & \textbf{0.61}      & 0.37            \\
                                                                       & WRanGAN        & \textbf{3.57} & 0.55               & \textbf{0.42}   \\ \hline
\end{tabular}
\caption{\textbf{WRanGAN} model quality evaluation performed using FID, Precision, and Recall metrics for two domains, FFHQ and LSUN Church. The best results for each domain and metric are highlighted in bold.}
\label{tab:fid_prec_rec}
\end{table}

\subsection{Weight randomization}
\label{sec:randomization}

The idea of randomizing the model was inspired by Bayesian GAN \cite{saatciwilson}, where the generator and discriminator networks both assumed to have some distribution over their internal parameters. However, randomizing the entire network is computationally expensive due to the increased number of parameters required for both training and generator parameters tuning during the inversion step. We have already illustrated how to perform inversion using appropriate regularization coefficients, and here, we present how to obtain such coefficients.

\paragraph{How many parameters to randomize?}



In \cite{alaluf2021hyperstyle}, experiments were conducted to determine the most effective parameters to be changed in the generator. It was decided to limit the randomization to the last few convolutional layers, excluding the toRGB layers, and to keep the discriminator architecture unchanged. To determine the appropriate number of layers for randomization, a grid search was conducted over $N = 4,6,8$ and different equal regularization coefficients for the \textbf{Simple Weight Tune} method. The results of this search are presented in Figure~\ref{fig:inv_layer_reg}, and the computational costs are shown in Table~\ref{tab:comp_n_rand_layers}. It was determined that randomizing only the last $N = 6$ convolutional layers yielded the best results with a minimal increase in computational costs.

\paragraph{How to train?}
To train the WRanGAN model, a pre-trained model was used to initialize the mean value of the model parameters $\mu_\theta = \theta_{G,0}$. Standard deviation was then added to each randomized parameter with value equal one. The generator and discriminator were trained together to reach the global optima, as outlined in Algorithm~\ref{alg:GAN_training}.

\section{Experiments}
\label{sec:test}

This section presents the results of the evaluation of the proposed \textbf{WRanGAN} model. Below are presented the details of conducted experiments: datasets, baselines, and hyperparameters.


\begin{figure*}[t]
   \centering
    \includegraphics[width=0.85\linewidth]{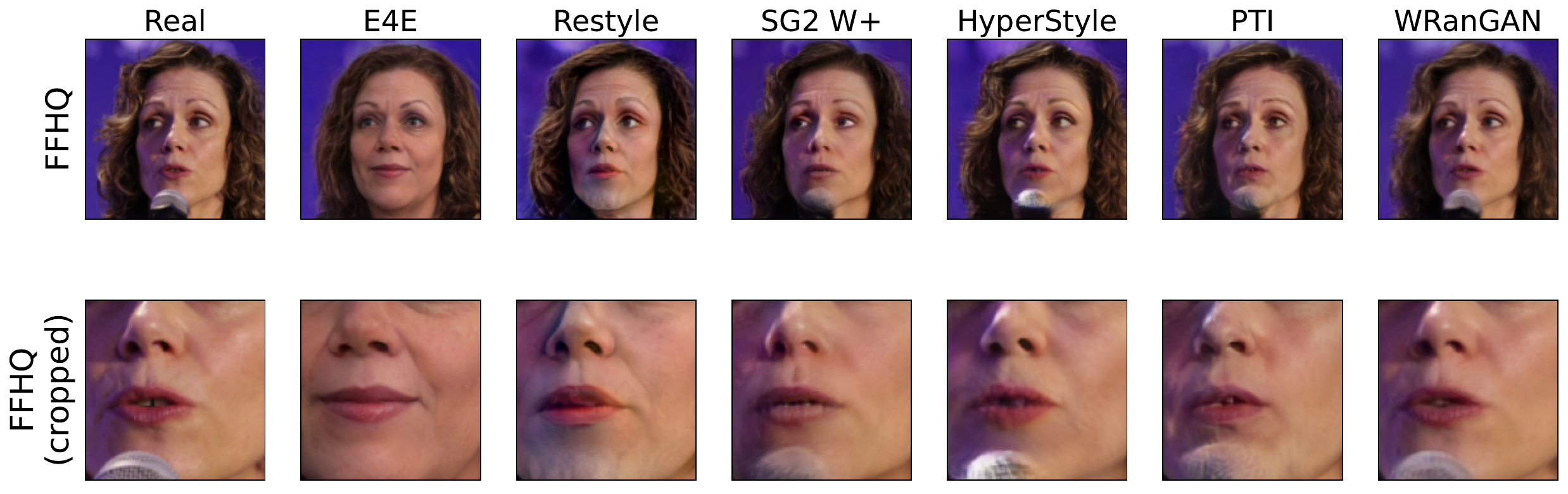}
    \caption{Qualitative evaluation of \textbf{WRanGAN} inversion results compared to ones produced by StyleGAN 2 using various approaches for FFHQ domain. For each reconstruction provided zoomed version (interesting regions were cropped) to see the difference in details completely.}
    \label{fig:rec_comp_ffhq}
\end{figure*}

\begin{figure*}[t]
   \centering
    \includegraphics[width=0.85\linewidth]{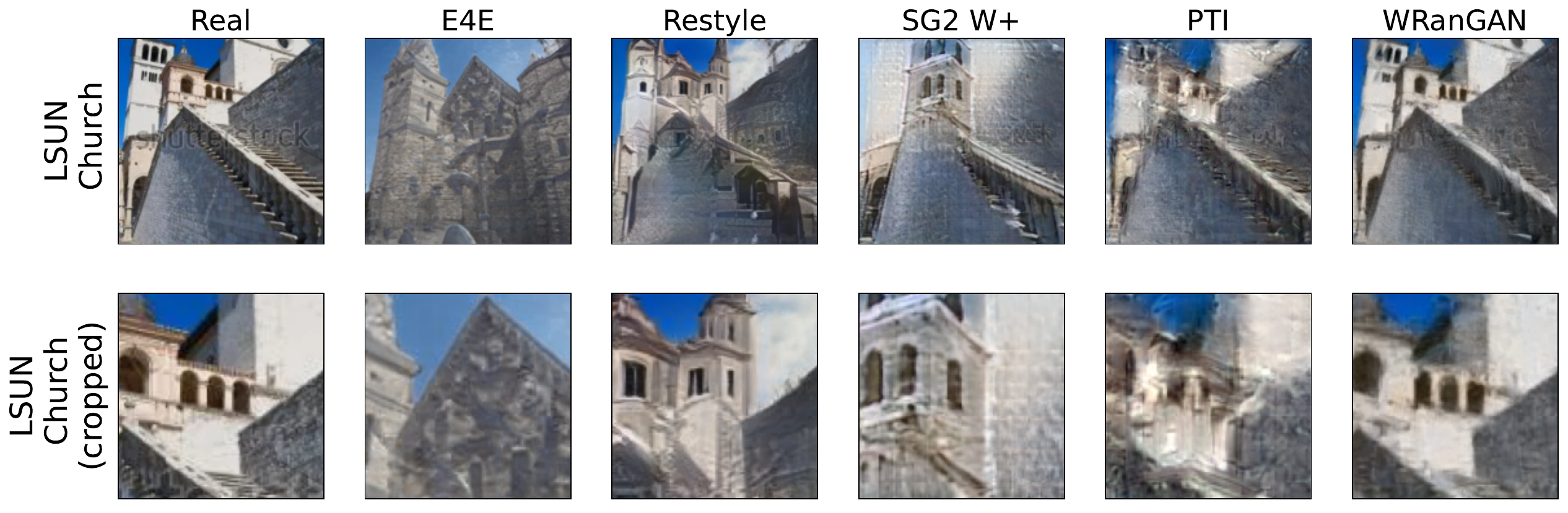}
    \caption{Qualitative evaluation of \textbf{WRanGAN} inversion results compared to ones produced by StyleGAN 2 using various approaches for LSUN Church domain. For each reconstruction provided zoomed version (interesting regions were cropped) to see the difference in details completely.}
    \label{fig:rec_comp_lsun}
\end{figure*}

\paragraph{Technical details.} 
    

    

\begin{itemize}
    \item \textbf{Models:} We used the StyleGAN 2 \cite{karras2020analyzing} model as a basis, with pre-trained models and base code for implementation taken from an open resource\footnote{https://github.com/rosinality/stylegan2-pytorch}.
    
    \item \textbf{Datasets:} We trained using the Flickr-Faces-HQ Dataset (FFHQ) \cite{karras2019stylebased} with pictures resized to resolution $256$x$256$, and LSUN Churches \cite{yu15lsun} with pictures center-cropped and resized to $256$x$256$. We randomly sampled 1000 images from both datasets for testing.

    \item \textbf{\textbf{WRanGAN} training details:} We used standard parameters for StyleGAN 2, and trained on 2 GPUs with a batch size of $8$ for $200k$ iterations.
    
    \item \textbf{\textbf{WRanGAN} inversion details:} For the encoder $E$ in Algorithm~\ref{alg:GAN_inversion}, we used the architecture proposed by \cite{Tov2021Designing} and trained with default parameters. We used the Adam optimizer with a learning rate of $lr= 10^{-3}$, and the number of iterations needed for convergence was set to $500$. The randomization parameter was initialized with the value $\epsilon = 10^{-4}$, and we used a regularization coefficient of $\alpha_{\mathrm{reg}} = 10^{-4}$.
\end{itemize}

Experiments were conducted on 4 Tesla V100-SXM2 GPUs with 16 GB of memory.



\subsection{\textbf{WRanGAN} model evaluation}



 We evaluated the performance of our WRanGAN model by running several metrics such as FID, Precision, and Recall \cite{Kynkaanniemi2019} and comparing our results with those produced by the StyleGAN 2 model (see Table~\ref{tab:fid_prec_rec}). WRanGAN showed an improvement in the Recall metric for both data domains, which suggests that the generator is more likely to reproduce particular real images. However, we observed a slight decrease in the Precision metric. For further details on the randomized parameters of the model and their effect on the generated images, please refer to Appendix A.

\subsection{Inversion quality assessment}

 Evaluating the quality of inversion, various encoder based approaches such as e4e \cite{Tov2021Designing}, ReStyle \cite{alaluf2021restyle}, and HyperStyle\cite{alaluf2021hyperstyle}, as well as optimization based approaches like SG2 W+ \cite{karras2020analyzing} and PTI \cite{roich2021pivotal}, were taken into consideration. Standard metrics including mean squared error (MSE), LPIPS\cite{zhang2018perceptual} with the VGG and Alex feature network, and MS-SSIM \cite{msssim} were used for assessment. As summarized in Table~\ref{tab:rec_comp}, the WRanGAN model proposed surpasses all the other methods applied to StyleGAN 2 on most metrics. Not only does it achieve the lowest distortion, its computational efficiency is also much higher compared to PTI, as the tuning procedure requires optimizing 4 times fewer parameters. This translates to 600 megabytes less GPU memory required during image inversion, making it more practical to use. Additionally, calculations are 1.5 and 2 times faster for FFHQ and LSUN Church domain respectively. Visualization of the results in Figure~\ref{fig:rec_comp_ffhq} and Figure~\ref{fig:rec_comp_lsun} further illustrate how the improvement in reconstruction affects the image, with the WRanGAN approach being able to reproduce unique details such as bangs, outline of the eyes, and small church windows. More detailed visualizations and an investigation of the distribution of randomized parameters for real mapped images can be found in Appendices B and C respectively.


\begin{table}[t]
\centering
\begin{tabular}{c|c|c|cc}
\hline
\textbf{Domain}                                                        & \textbf{Model}              & \textbf{Method}                                             & \textbf{FID}& \textbf{KID} \\ \hline
\multirow{3}{*}{FFHQ}                                                  & \multirow{2}{*}{StyleGAN 2} & Restyle                                                     & 244          & 0.25          \\
                                                                       &                             & PTI                                                         & 222          & 0.225          \\ \cline{2-5} 
                                                                       & WRanGAN                     & \begin{tabular}[c]{@{}c@{}}WRanGAN\\ inversion\end{tabular} & 225          & 0.232          \\ \hline
\multirow{3}{*}{\begin{tabular}[c]{@{}c@{}}LSUN\\ Church\end{tabular}} & \multirow{2}{*}{StyleGAN 2} & Restyle                                                     & 139          & 0.128          \\
                                                                       &                             & PTI                                                         & 133          & 0.134          \\ \cline{2-5} 
                                                                       & WRanGAN                     & \begin{tabular}[c]{@{}c@{}}WRanGAN\\ inversion\end{tabular} & 144          & 0.144          \\ \hline
\end{tabular}
\caption{Model corruption evaluation. The two best methods for StyleGAN 2 were taken for comparison:\textbf{PTI} and Restyle. The lower values are better for each metric.}
\label{tab:fid_kid_comp}
\end{table}

\begin{figure*}[t]
   \centering
    \includegraphics[width=0.95\linewidth]{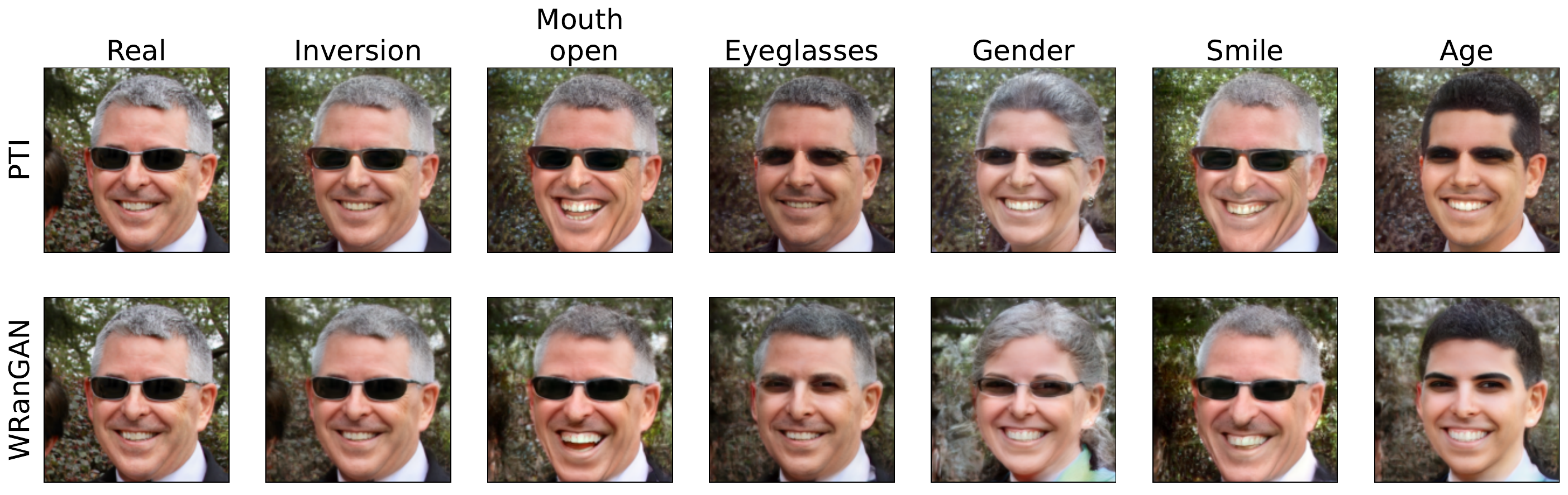}
    \caption{Qualitative evaluation of \textbf{WRanGAN} editing quality compared to \textbf{PTI} approach applied over StyleGAN 2 model.}
    \label{fig:editing_comp_all}
\end{figure*}

\begin{figure*}[t]
   \centering
    \includegraphics[width=0.95\linewidth]{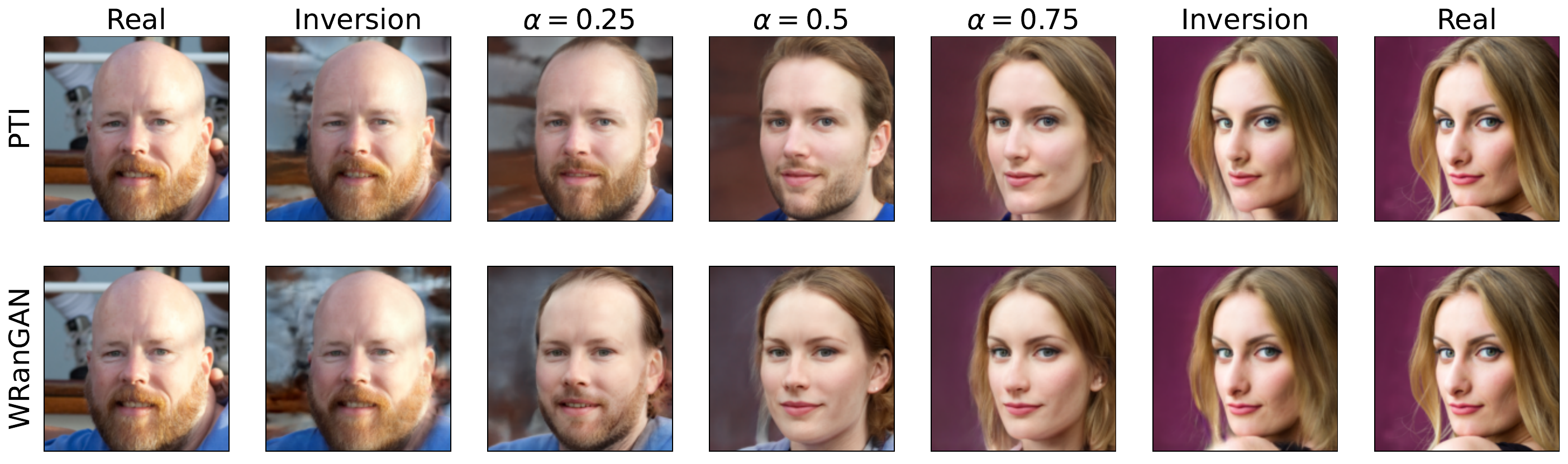}
    \caption{Qualitative evaluation of \textbf{WRanGAN} interpolation quality compared to \textbf{PTI} approach applied over StyleGAN 2 model. Here $\alpha$ denotes interpolation step.}
    \label{fig:interpolation_comp_all}
\end{figure*}

\subsection{Model corruption evaluation}
Initially, we mentioned the effect of tuning the model parameters on the ability to generate realistic images. In this part, we performed an assessment of model corruption of tuned models using FID and KID metrics. The comparison performed with two most efficient encoder and optimization based approaches: Restyle and \textbf{PTI}. The results for both domains are presented in Table~\ref{tab:fid_kid_comp}. The difference in FID and KID values between the proposed \textbf{WRanGAN} and StyleGAN 2 inversion approaches does not exceed $11$ for both domains, which is much smaller than $24$ (quality drop demonstrated by the \textbf{Simple Weight Tune} regularization strategy). As a result, we concluded that \textbf{WRanGAN} preserves ability generate realistic images after tuning. 


\begin{table}[t]
\centering
\begin{tabular}{c|cc}
\hline
\textbf{Attribute}  & \textbf{StyleGAN 2} & \textbf{WRanGAN} \\ \hline
\textbf{Gender}     & 73.9                & \textbf{75.0}    \\
\textbf{Eyeglasses} & 99.8                & \textbf{99.9}    \\
\textbf{Smile}      & 99.5                & \textbf{99.8}    \\
\textbf{Age}        & \textbf{99.5}       & 99.4             \\
\textbf{Mouth open} & 98.2                & \textbf{98.5}    \\ \hline
\end{tabular}
\caption{Classification accuracy (\%) on separation boundaries in latent space with respect to different face attributes. The best results are highlighted in bold.}
\label{tab:classification_acc}
\end{table}

\subsection{Editing and interpolation quality assessment}


We conducted an experiment to confirm that the \textbf{WRanGAN} model possesses the same excellent property as the StyleGAN 2 model - for any binary attribute, there exists a hyperplane in latent space such that all samples from the same side have the same attribute \cite{shen2020interpreting}. To do this, we trained a classifier predicting the following attributes: Gender, Eyeglasses, Smile, Age, Open Mouth. We then constructed hyperplanes in the latent space corresponding to the selected attributes and evaluated their correctness, as shown in Table~\ref{tab:classification_acc}. The results of our experiment demonstrate that the \textbf{WRanGAN} model has slightly superior performance compared to the basic StyleGAN 2 model. This is also evident in the visualization presented in Figure~\ref{fig:editing_comp_all}, where it is easy to notice that the presence of glasses in the original image significantly affected the editing of attributes in the case of the \textbf{PTI} method. However, \textbf{WRanGAN} demonstrates an excellent performance. It's also noticeable from interpolation comparison presented in Figure~\ref{fig:interpolation_comp_all}. For more examples please refer to Appendix D.

\section{Conclusion}


We have proposed a randomized version of the StyleGAN 2 model, dubbed WRanGAN, which is able to learn the appropriate scaling (standard deviation) for each parameter defining the corresponding regularization coefficient. Our approach to GAN tuning using non-equal regularization coefficients demonstrated superior results in terms of distortion and computational efficiency compared to the most successful approach, pivotal tuning inversion. Moreover, it did not corrupt the model, allowing for image editing. We also showed that in the latent space of a randomized model, it is easy to construct a hyperplane corresponding to the standard image attributes for the FFHQ domain.

Our method requires less memory per image in the inversion process, making it easy to parallelize calculations. Additionally, the method is slightly dependent on the network architecture enabling to transfer to other structures such as StyleGAN 3 \cite{Karras2021}.

\section*{Ethical Statement}

The considered approach allows one to reproduce a photo of a real person with high accuracy. So, the user can edit photos of real people, which can lead to illegal actions.

{\small
\bibliographystyle{ieee_fullname}
\bibliography{egbib}
}

\newpage
.
\newpage
\appendix
\newpage
\section{Investigation of WRanGAN model randomization}
\begin{figure}[h!]
   \centering
    \includegraphics[width=1.0\linewidth]{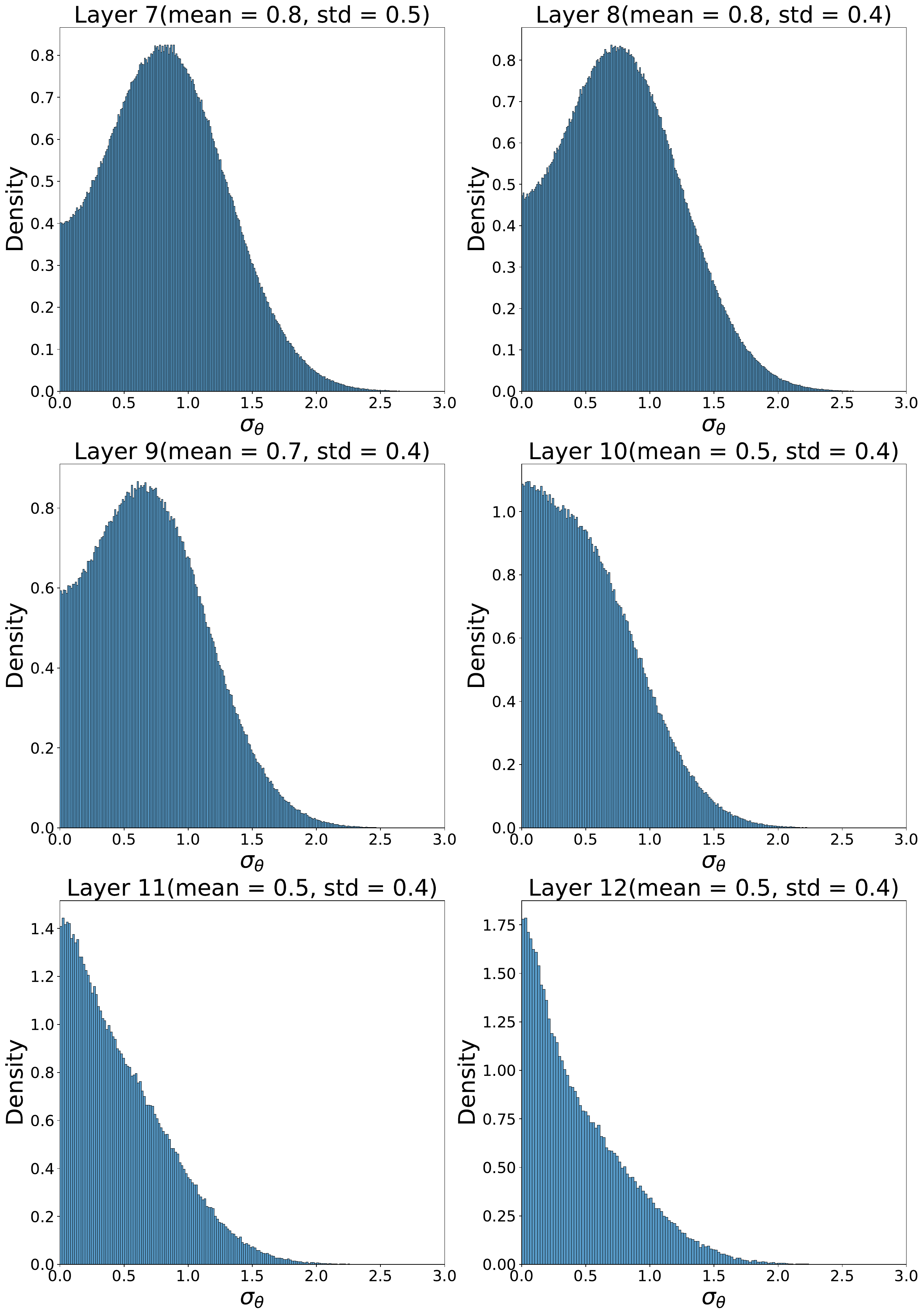}
    \caption{Dependence of distribution of randomized parameters variance with respect to index of convolutional layer for model trained on FFHQ dataset.}
    \label{fig:ffhq_variance}
\end{figure}
\begin{figure}[h!]
   \centering
    \includegraphics[width=1.0\linewidth]{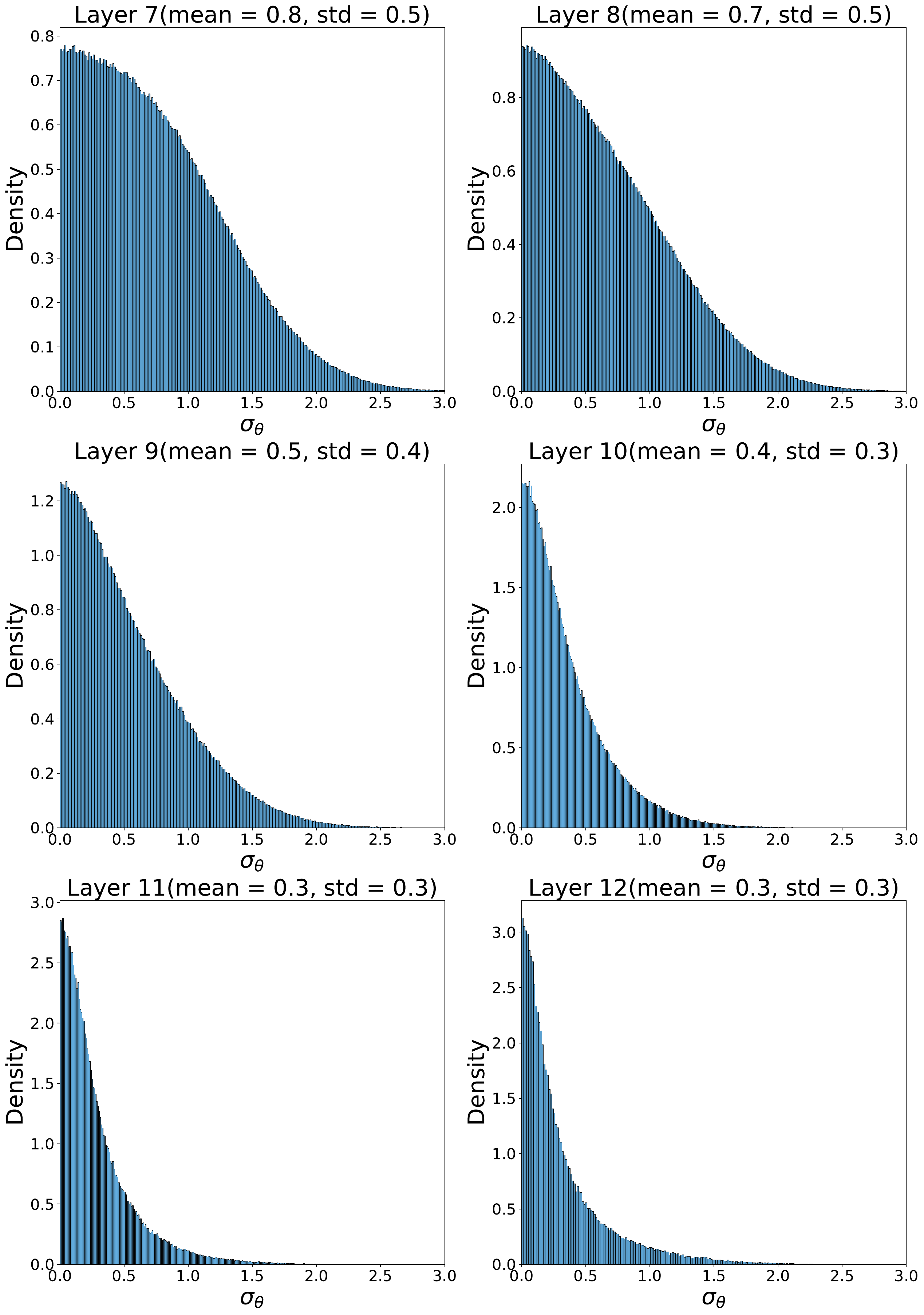}
    \caption{Dependence of distribution of randomized parameters variance with respect to index of convolutional layer for model trained on LSUN Church dataset.}
    \label{fig:church_variance}
\end{figure}
\begin{table}[h!]
\centering
\begin{tabular}{c|c}
\hline
\textbf{Layer index} & \textbf{Percentage} \\ \hline
7                    & 0.13 \%                \\
8                    & 0.16 \%                \\
9                    & 0.18 \%                \\
10                   & 0.29 \%                \\
11                   & 0.40 \%                \\
12                   & 0.52 \%                \\ \hline
\end{tabular}
\caption{Percentage of small variances ($\sigma_\theta < 10^{-3}$) for each randomized convolutional layer. }
\label{tab:small_variance}
\end{table}
\begin{figure*}[h!]
   \centering
    \includegraphics[width=1.0\linewidth]{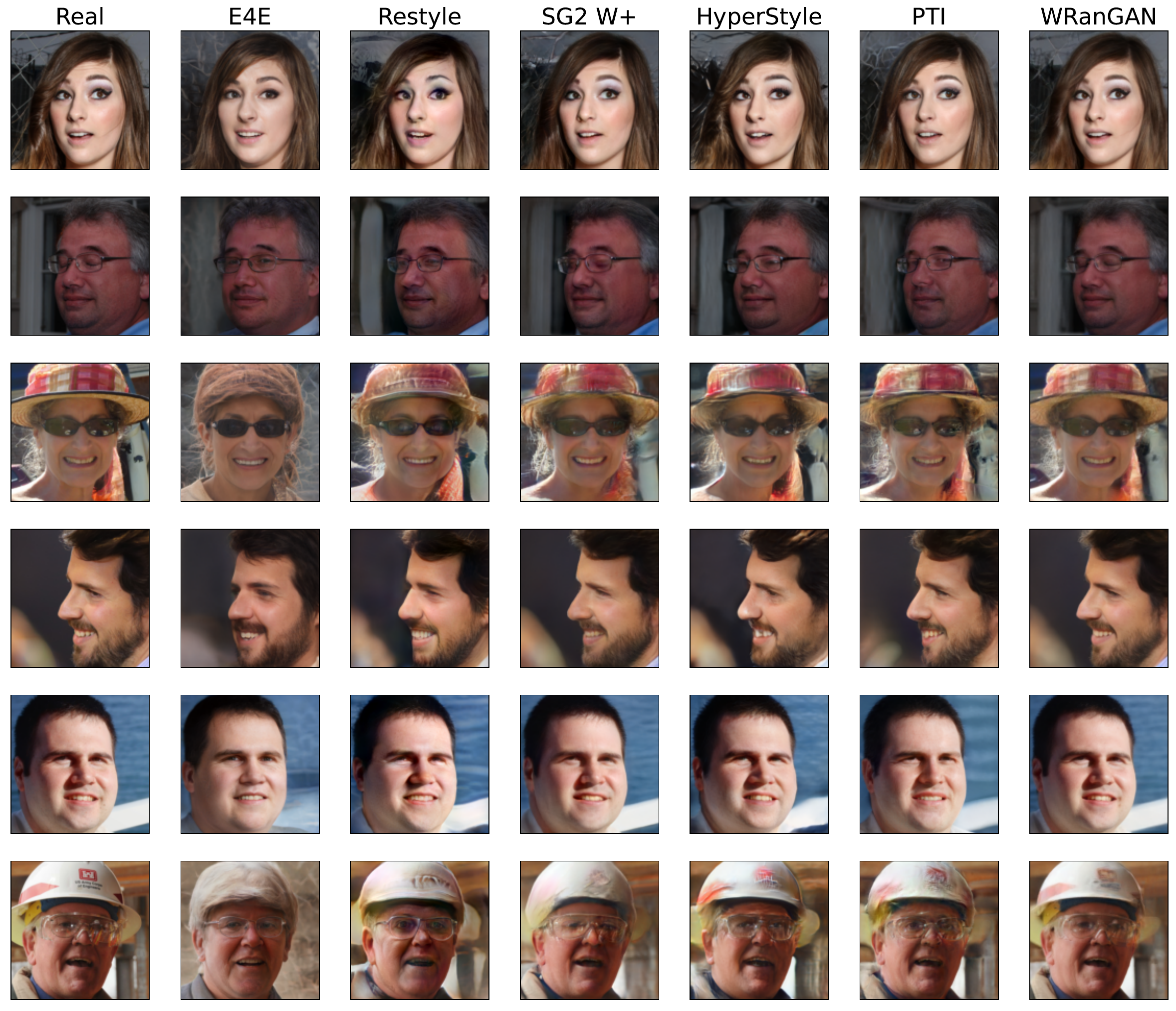}
    \caption{Qualitative reconstruction examples for images taken from FFHQ dataset.}
    \label{fig:add_rec_comp_ffhq}
\end{figure*}
\begin{figure*}[h!]
   \centering
    \includegraphics[width=1.0\linewidth]{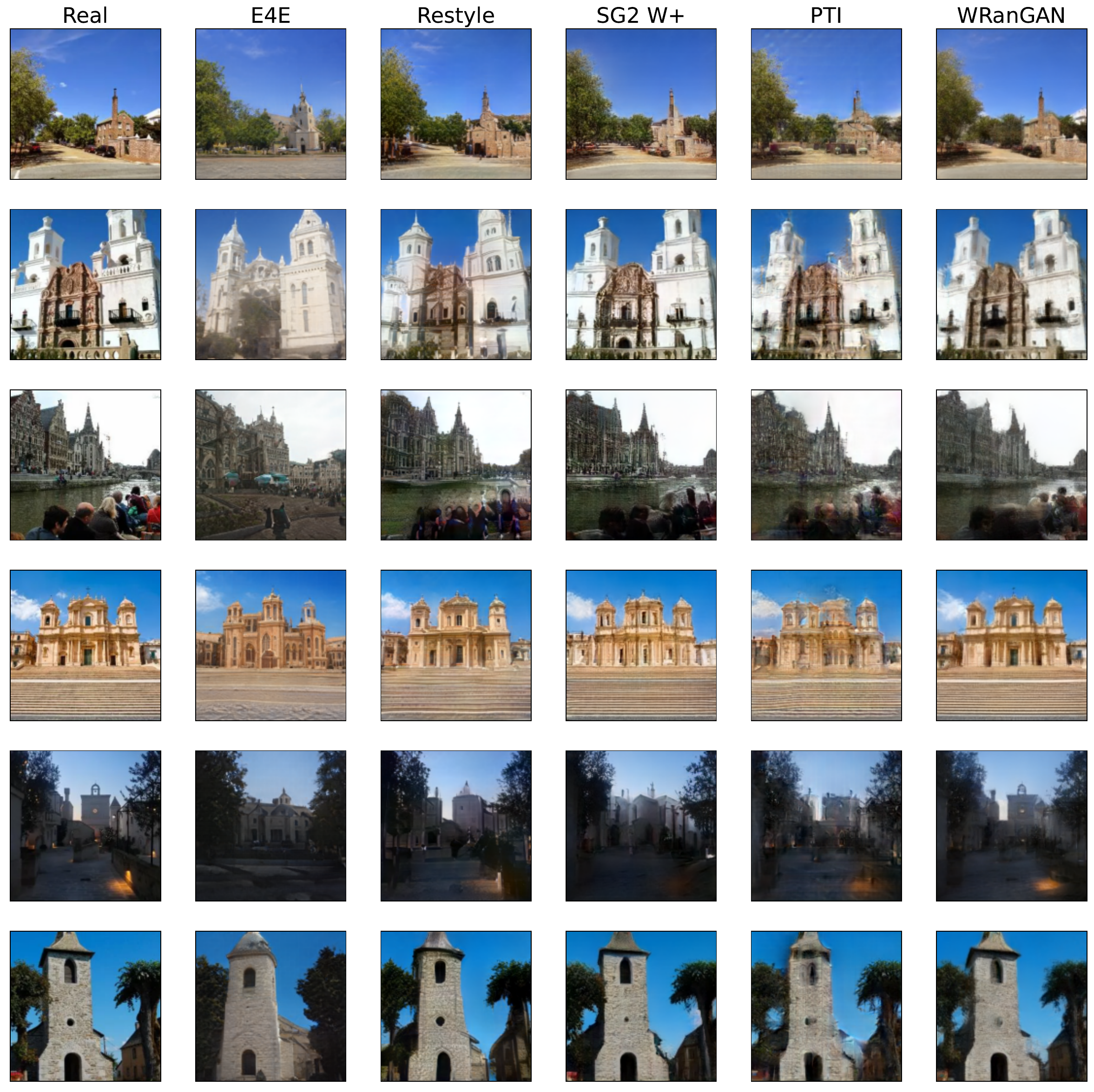}
    \caption{Qualitative reconstruction examples for images taken from LSUN Church dataset.}
    \label{fig:add_rec_comp_lsun}
\end{figure*}
 
\begin{table}[h!]
\centering
\begin{tabular}{c|c}
\hline
\textbf{Layer index} & \textbf{MSE} \\ \hline
7                    & 0.017                \\
8                    & 0.017                \\
9                    & 0.015                \\
10                   & 0.015                \\
11                   & 0.014                \\
12                   & 0.014                \\ \hline
\end{tabular}
\caption{Influence of each randomized layer on generator output.}
\label{tab:layer_influence}
\end{table}

The WRanGAN model is a type of generative model that optimizes two parameters - the mean and variance - in its learning process. It is hypothesized that a larger variance value allows for more changes to be made, which can be observed by examining the distributions of the variance values of the trained WRanGAN model. Figures~\ref{fig:ffhq_variance} and~\ref{fig:church_variance} present the distributions for the FFHQ and LSUN Church domains respectively. It can be seen that the distribution is shifted towards zero as the number of randomized layers increases. In order to verify this idea, a test was conducted to measure the number of parameters that have a variance close to zero ($\sigma_\theta < 10^{-3}$) in each layer. The results, as presented in Table~\ref{tab:small_variance}, indicate that the last layers have a greater effect on the model performance. To further investigate this idea, a procedure was conducted in which the parameters of different layers were changed and the impact of these changes on the output image was assessed by calculating the MSE metric (Table~\ref{tab:layer_influence}). The results confirm the hypothesis that the last layers have the greatest influence. 
\begin{figure*}[h!]
   \centering
    \includegraphics[width=1.0\linewidth]{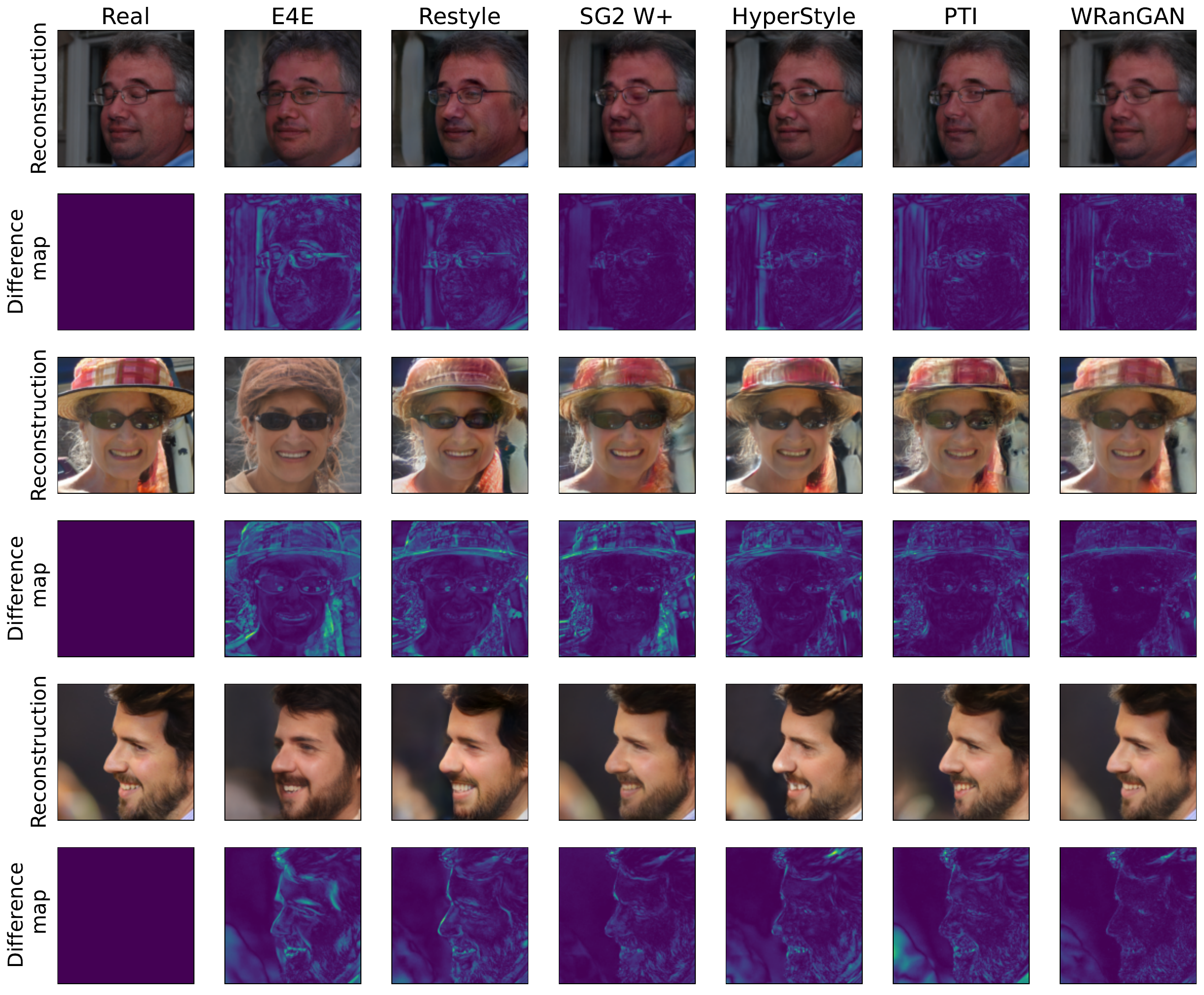}
    \caption{Qualitative reconstruction examples for FFHQ dataset with difference map, which represents pixel wise difference between reconstructed and original images. Map colors range from purple to yellow - from exact match to maximal difference.}
    \label{fig:heat_map_ffhq}
\end{figure*}
\begin{figure*}[h!]
   \centering
    \includegraphics[width=1.0\linewidth]{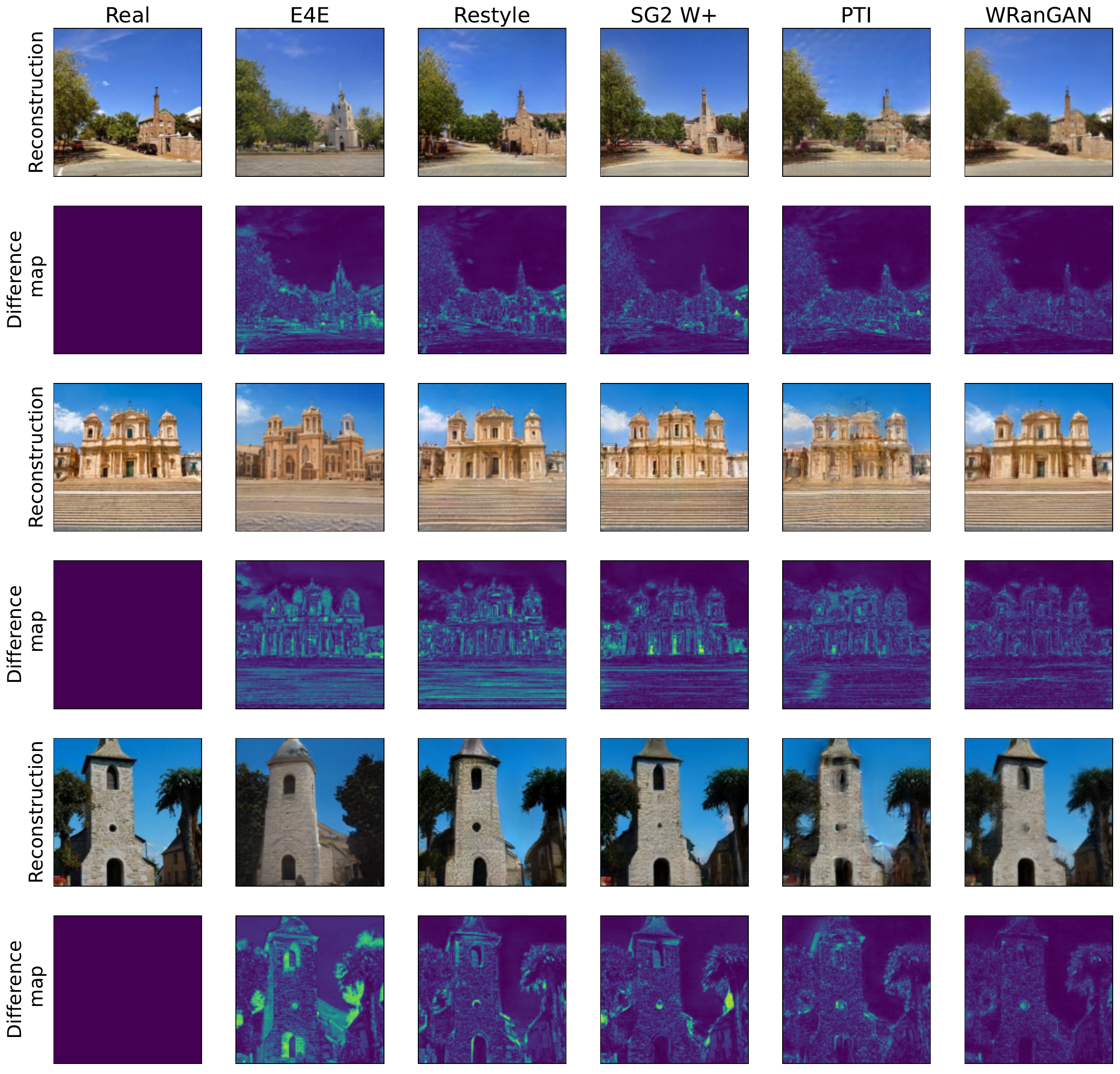}
    \caption{Qualitative reconstruction examples for LSUN Church dataset with difference map, which represents pixel wise difference between reconstructed and original images. Map colors range from purple to yellow - from exact match to maximal difference.}
    \label{fig:heat_map_church}
\end{figure*}
\section{Additional qualitative results on reconstruction}

The proposed approach of WRanGAN demonstrates a clear advantage in terms of reproducing unique details, as demonstrated in the additional examples of Figure~\ref{fig:add_rec_comp_ffhq} and Figure~\ref{fig:add_rec_comp_lsun}. Difference maps in Figure~\ref{fig:heat_map_ffhq} and Figure~\ref{fig:heat_map_church} more clearly show the areas of the image that differ from the original, where the WRanGAN model has more accurately reproduced the unique details of facial skin tones, wrinkles, clothing elements, complex hairstyles, and background elements, such as the windows, friezes, pilasters, and clock in the case of churches. However, some unique details are not completely restored even with the use of the new approach.

\begin{figure}[h!]
   \centering
    \includegraphics[width=0.81\linewidth]{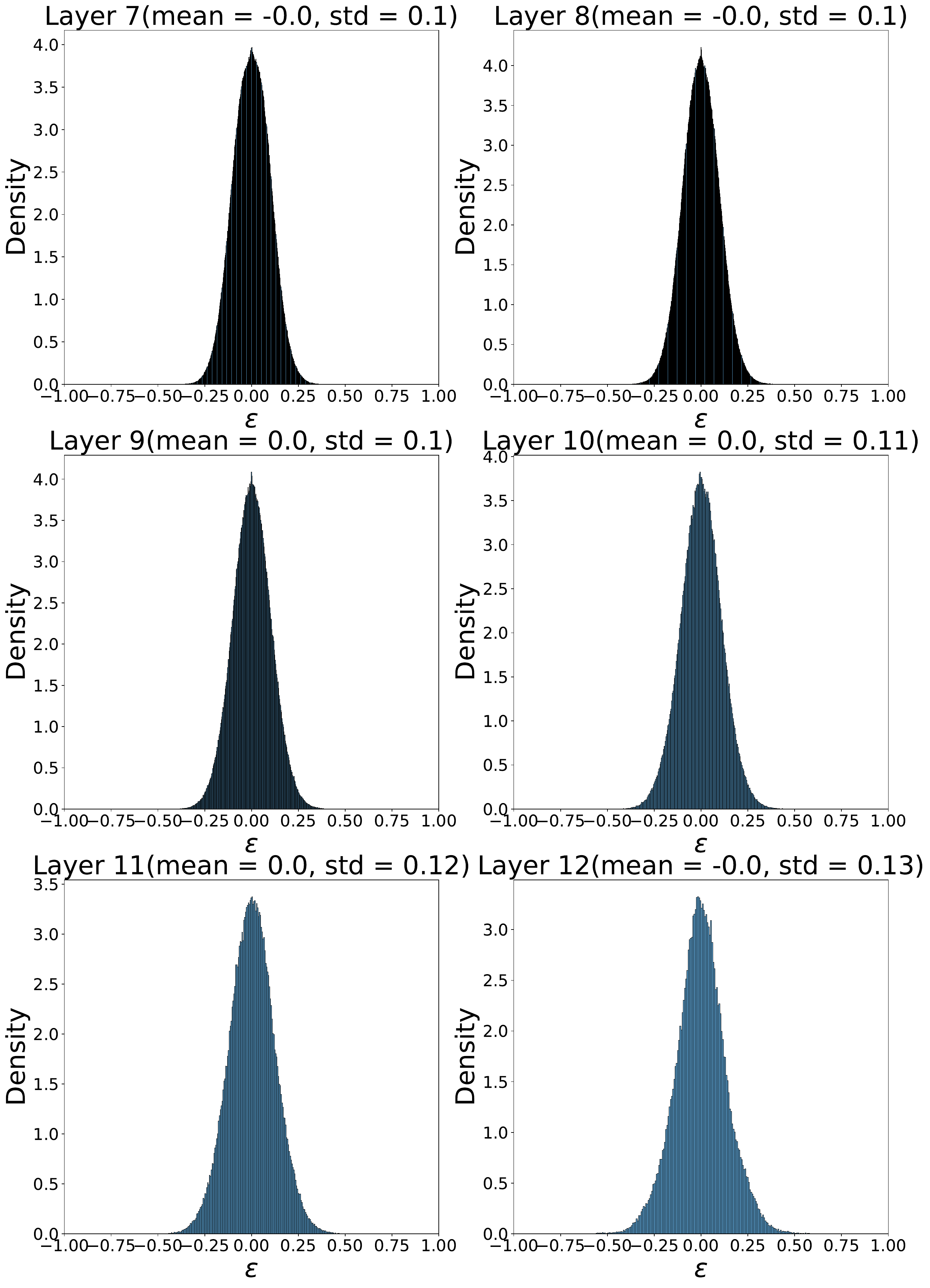}
    \caption{Distribution of parameter $\epsilon$ among randomized layers for image taken from FFHQ.}
    \label{fig:eps_dist_ffhq_1}
\end{figure}
\begin{figure}[h!]
   \centering
    \includegraphics[width=0.81\linewidth]{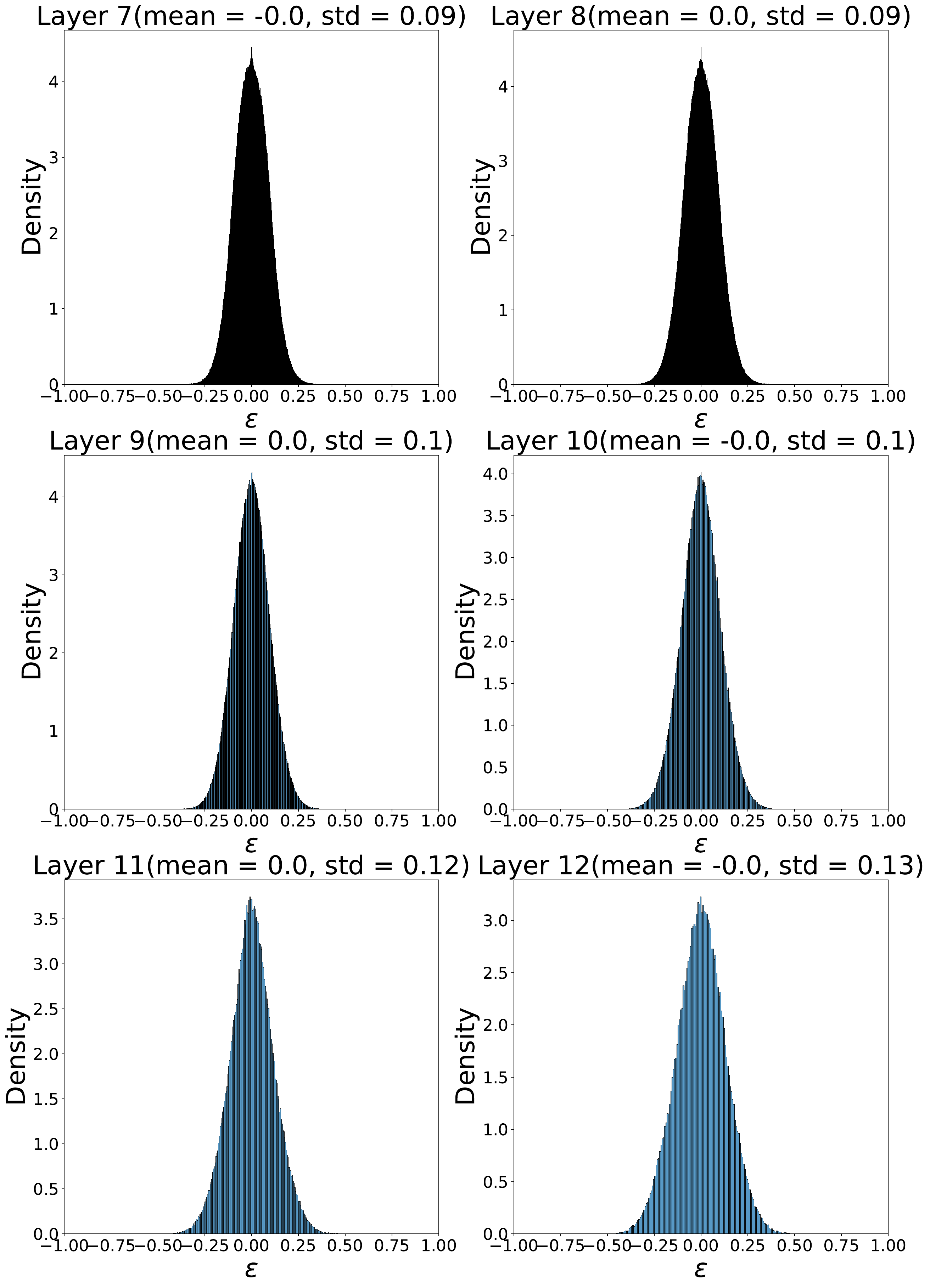}
    \caption{Distribution of parameter $\epsilon$ among randomized layers for image taken from FFHQ.}
    \label{fig:eps_dist_ffhq_2}
\end{figure}
\begin{figure}[h!]
   \centering
    \includegraphics[width=0.81\linewidth]{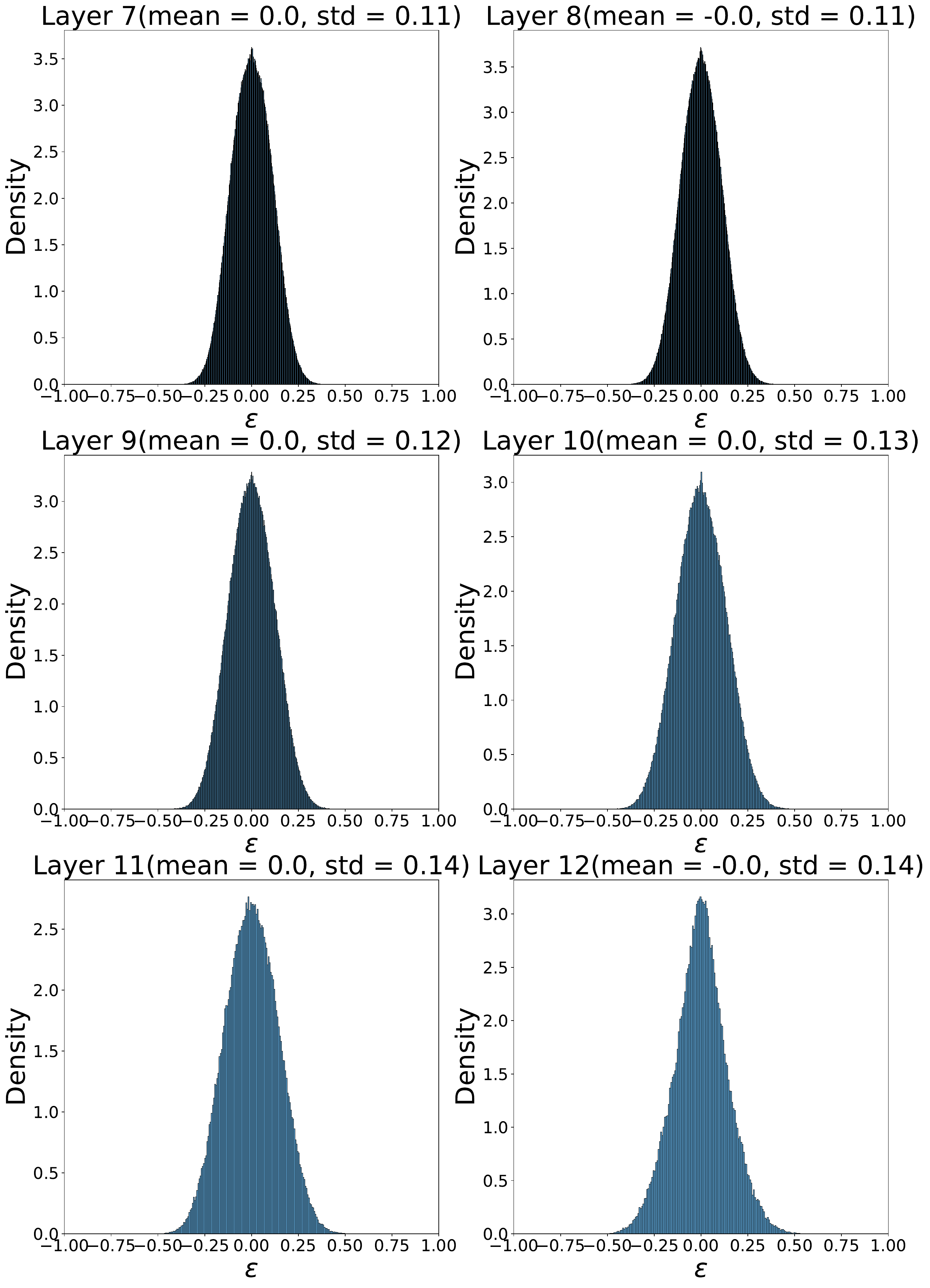}
    \caption{Distribution of parameter $\epsilon$ among randomized layers for image taken from FFHQ.}
    \label{fig:eps_dist_ffhq_3}
\end{figure}
\begin{figure}[h!]
   \centering
    \includegraphics[width=0.81\linewidth]{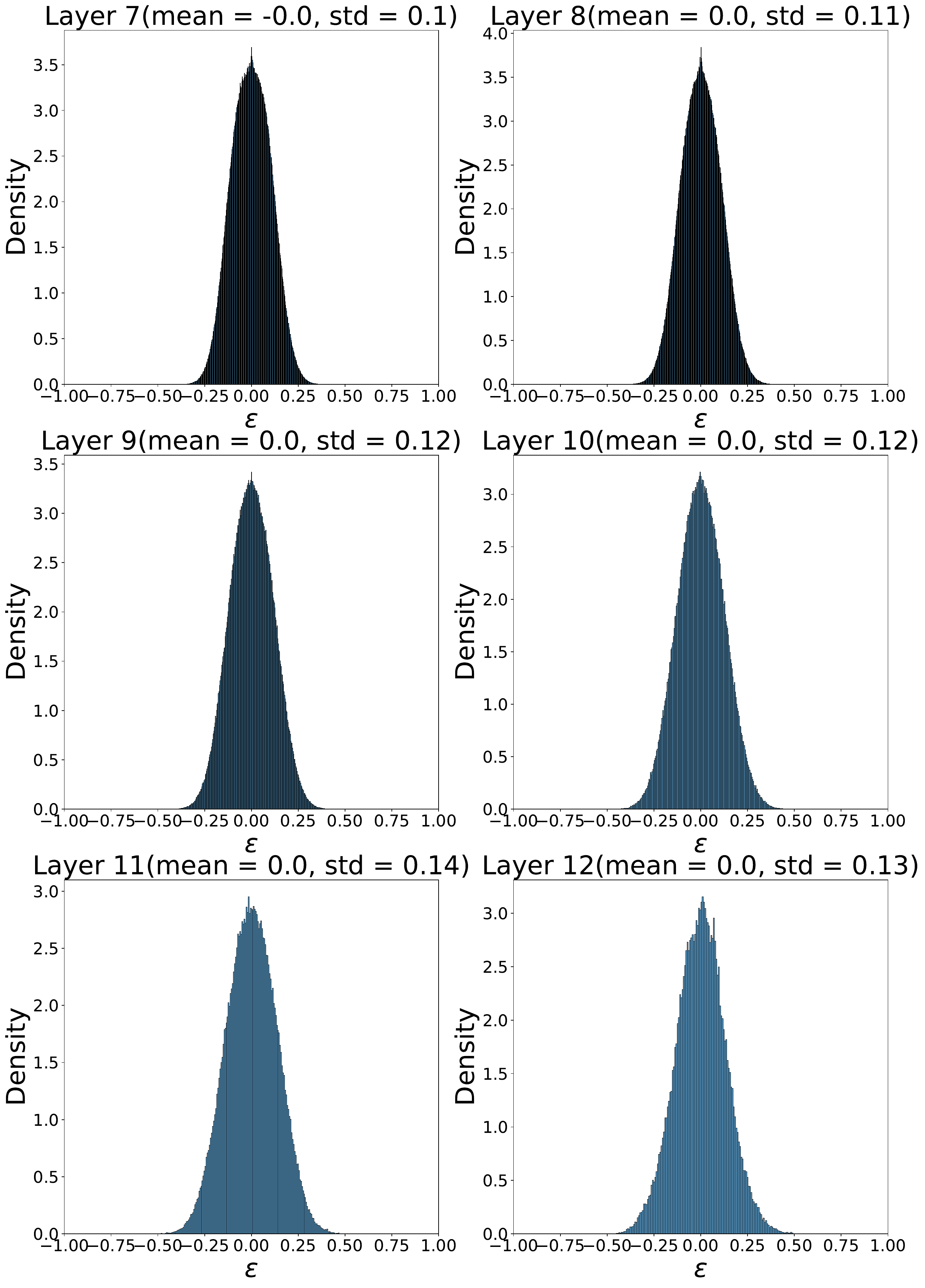}
    \caption{Distribution of parameter $\epsilon$ among randomized layers for image taken from FFHQ.}
    \label{fig:eps_dist_ffhq_4}
\end{figure}
\begin{figure}[h!]
   \centering
    \includegraphics[width=0.81\linewidth]{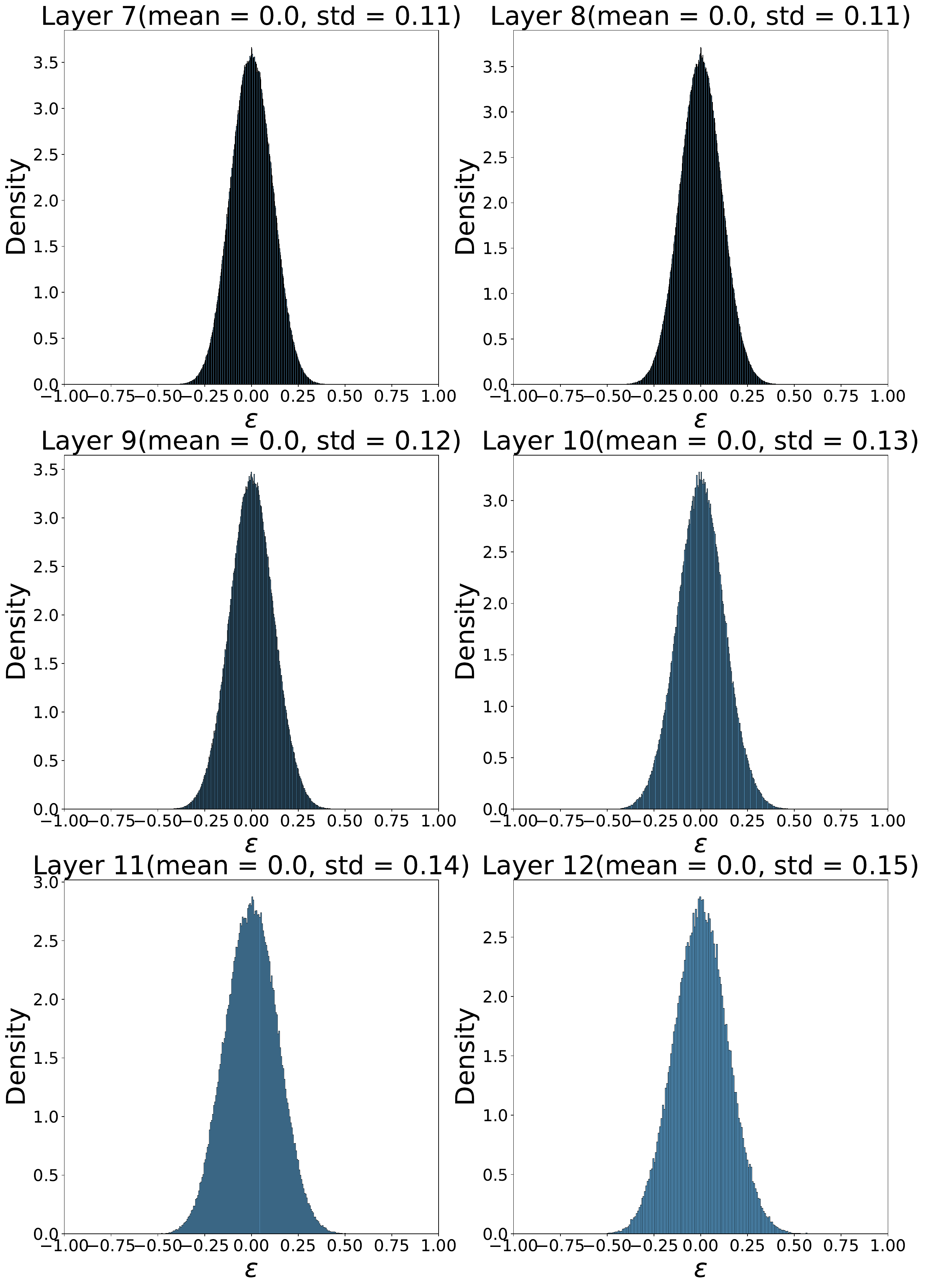}
    \caption{Distribution of parameter $\epsilon$ among randomized layers for image taken from FFHQ.}
    \label{fig:eps_dist_ffhq_5}
\end{figure}

\begin{figure}[h!]
   \centering
    \includegraphics[width=0.81\linewidth]{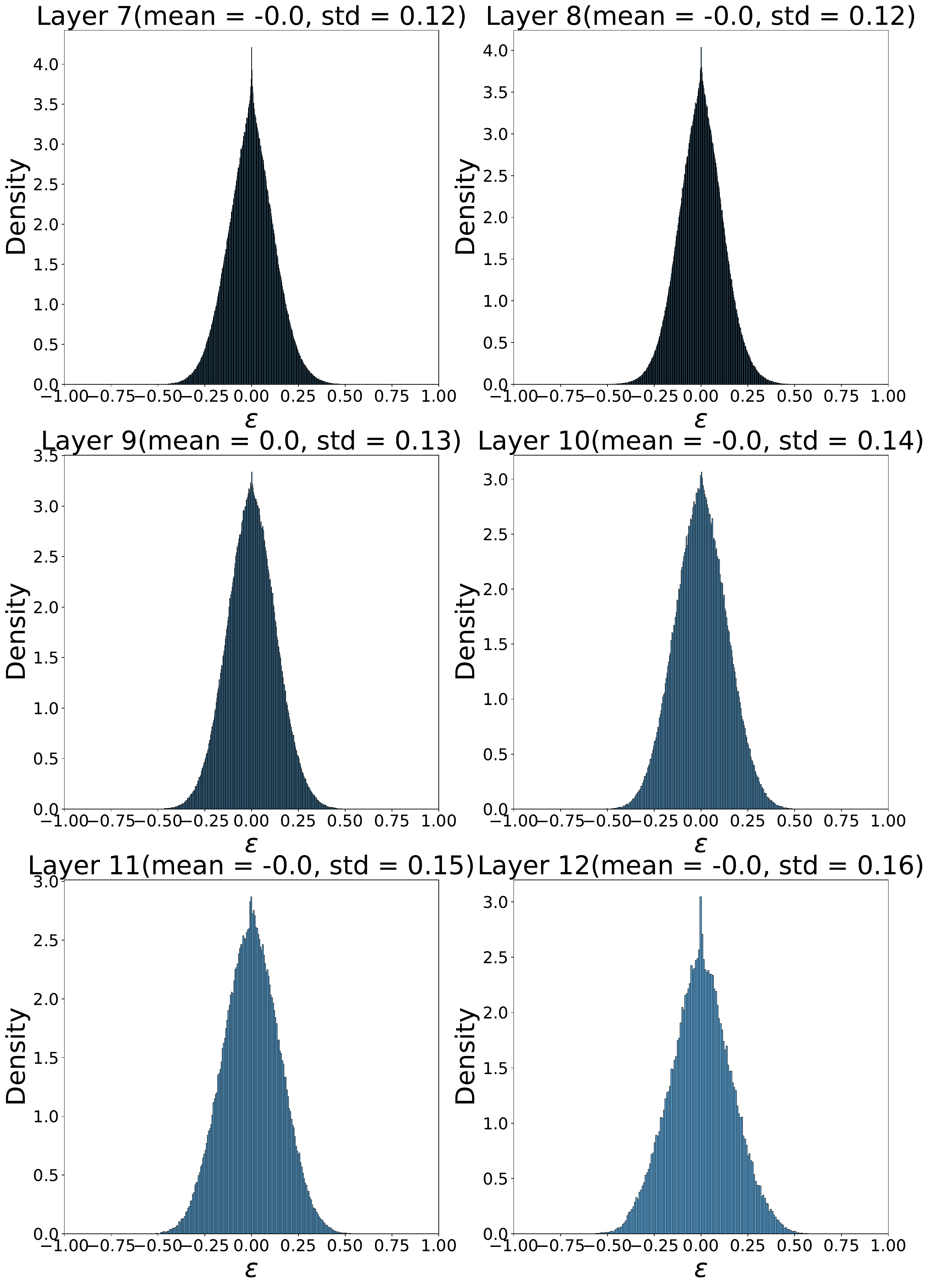}
    \caption{Distribution of parameter $\epsilon$ among randomized layers for image taken from LSUN Church.}
    \label{fig:eps_dist_church_1}
\end{figure}
\begin{figure}[h!]
   \centering
    \includegraphics[width=0.81\linewidth]{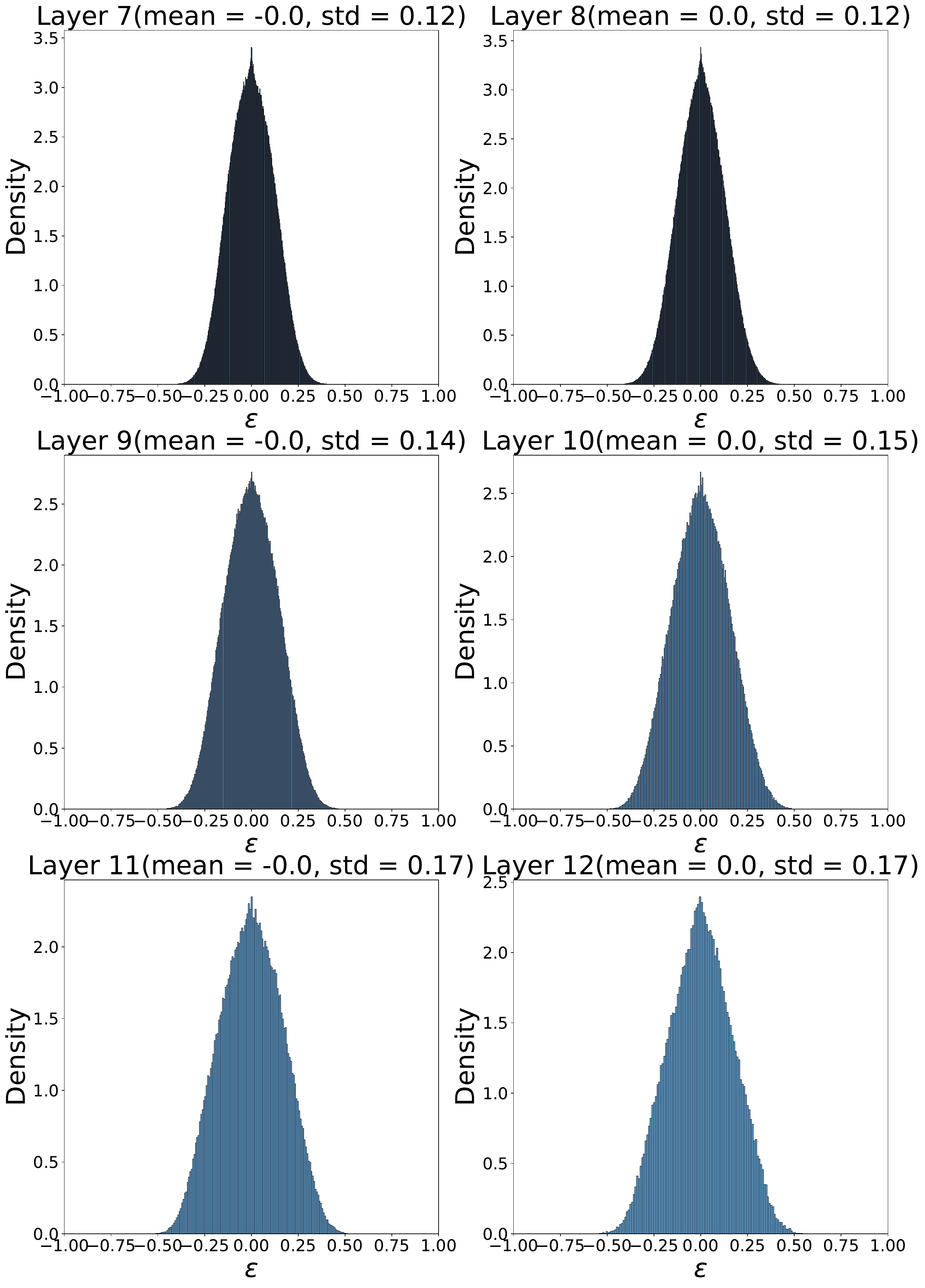}
    \caption{Distribution of parameter $\epsilon$ among randomized layers for image taken from LSUN Church.}
    \label{fig:eps_dist_church_2}
\end{figure}
\begin{figure}[h!]
   \centering
    \includegraphics[width=0.81\linewidth]{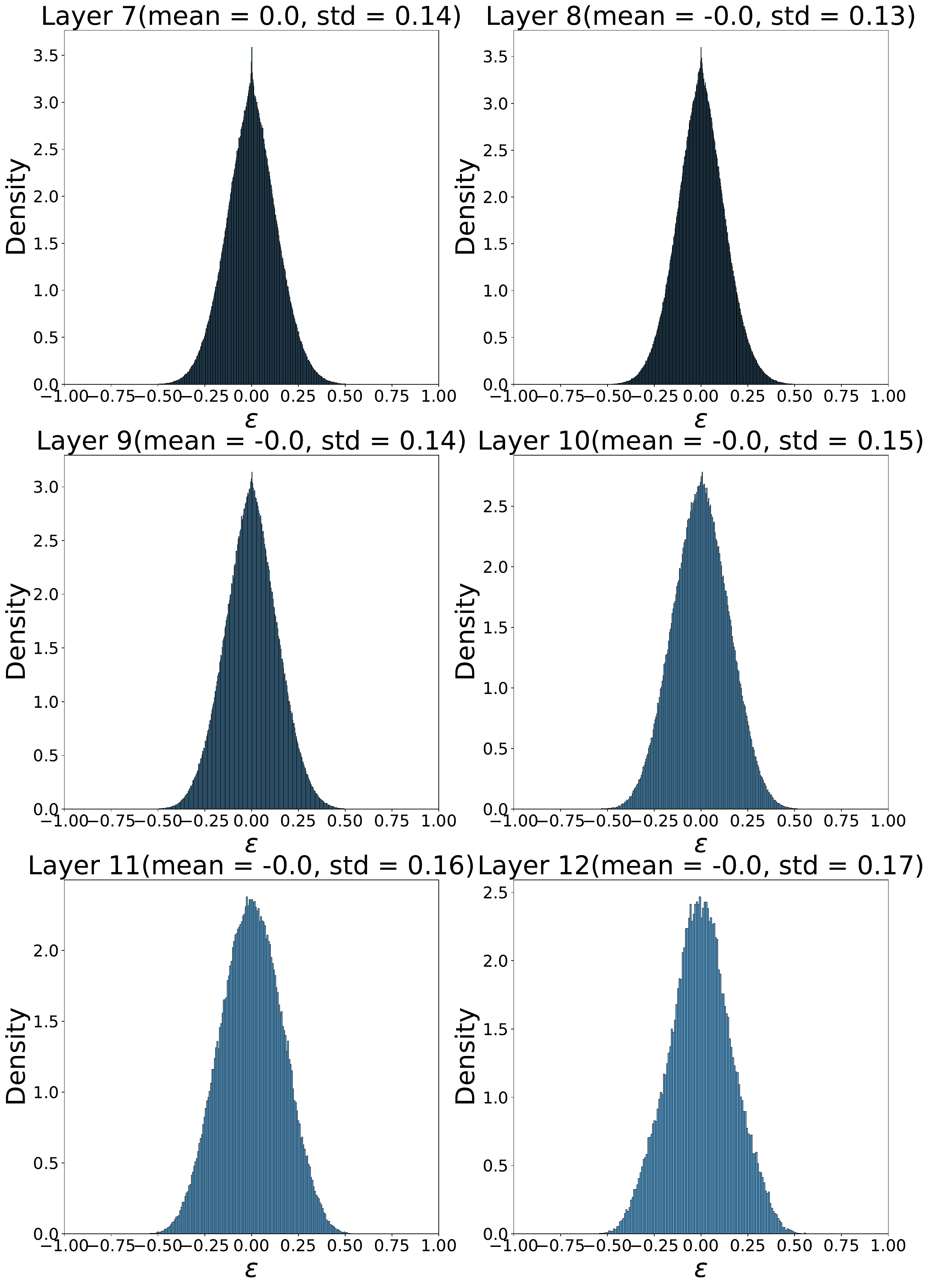}
    \caption{Distribution of parameter $\epsilon$ among randomized layers for image taken from LSUN Church.}
    \label{fig:eps_dist_church_3}
\end{figure}
\begin{figure}[h!]
   \centering
    \includegraphics[width=0.81\linewidth]{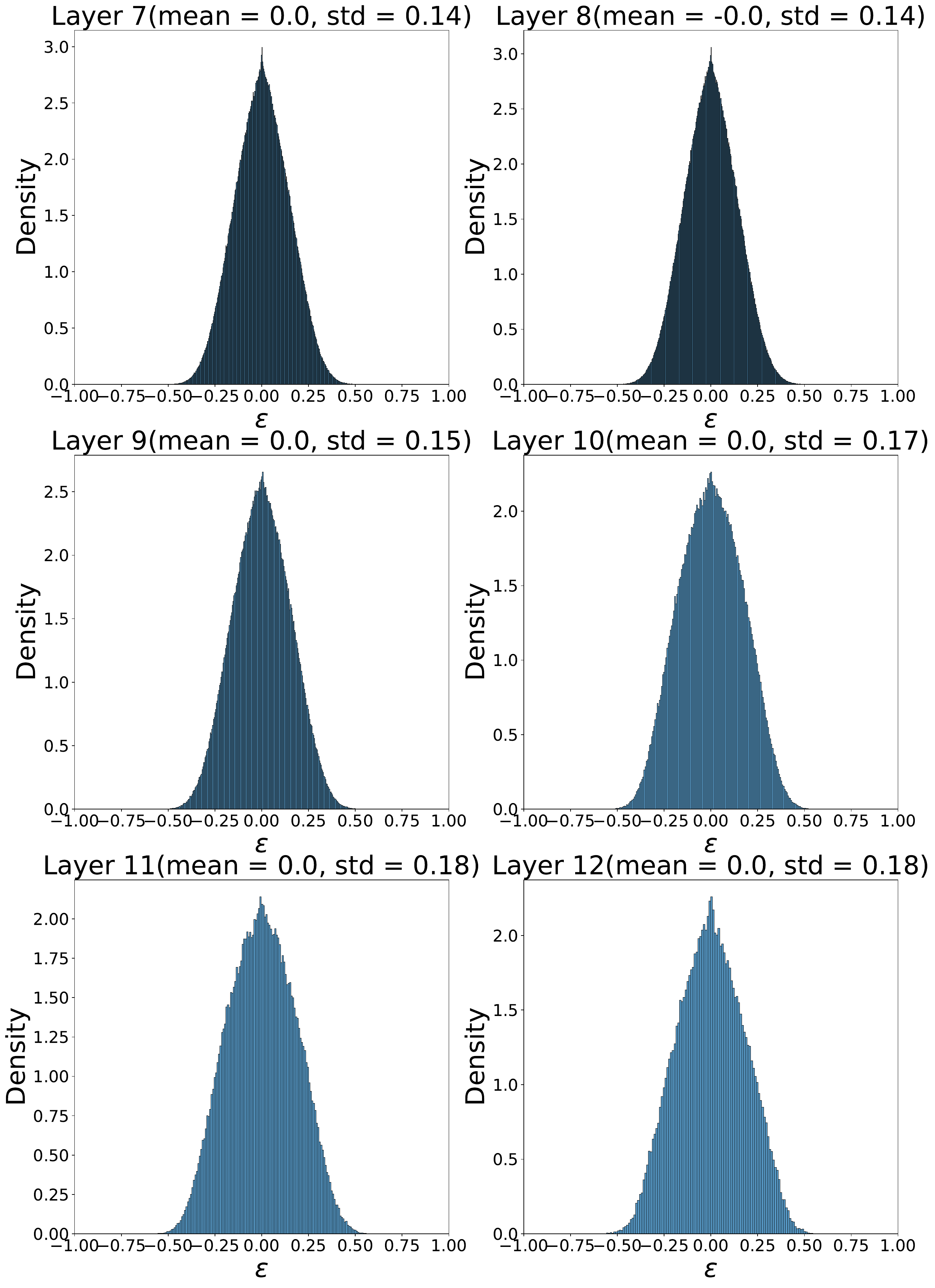}
    \caption{Distribution of parameter $\epsilon$ among randomized layers for image taken from LSUN Church.}
    \label{fig:eps_dist_church_4}
\end{figure}
\begin{figure}[h!]
   \centering
    \includegraphics[width=0.81\linewidth]{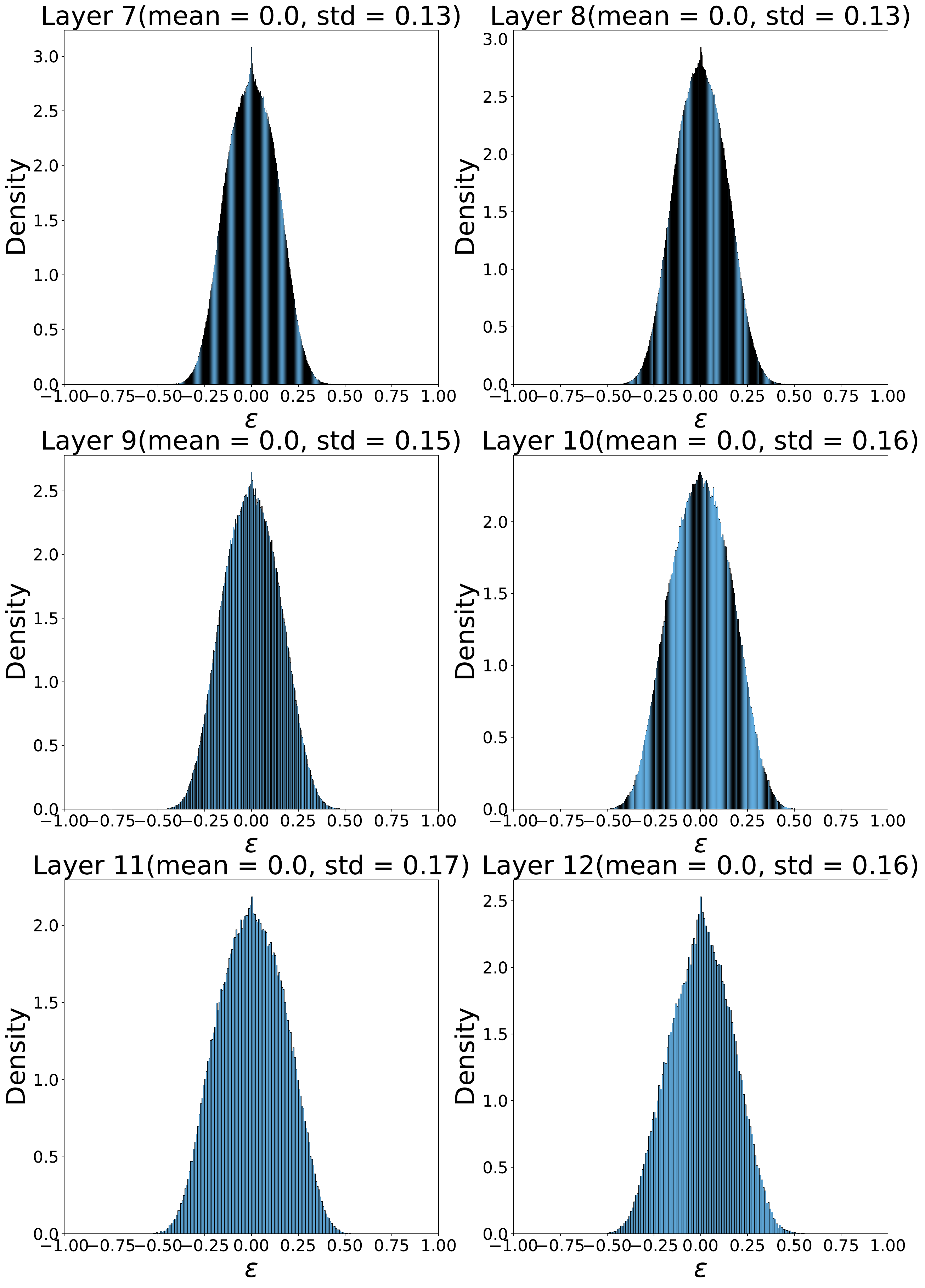}
    \caption{Distribution of parameter $\epsilon$ among randomized layers for image taken from LSUN Church.}
    \label{fig:eps_dist_church_5}
\end{figure}

\section{Investigation of inversion procedure results}

The proposed methodology utilizes a reparametrization trick to work with randomized parameters. Specifically, the generator parameter $\theta_g$ is represented as $\theta_g = \mu_\theta + \epsilon \sigma_\theta$, where $\epsilon$ is sampled from a normal distribution. Figures~\ref{fig:eps_dist_ffhq_1}, \ref{fig:eps_dist_ffhq_2}, \ref{fig:eps_dist_ffhq_3}, \ref{fig:eps_dist_ffhq_4}, \ref{fig:eps_dist_ffhq_5}, and Figures~\ref{fig:eps_dist_church_1}, \ref{fig:eps_dist_church_2}, \ref{fig:eps_dist_church_3}, \ref{fig:eps_dist_church_4}, \ref{fig:eps_dist_church_5}, provide examples of the resulting distribution of $\epsilon$ for images taken from FFHQ and LSUN Church respectively. In general, the parameter is close to normally distributed, albeit with a variance that is 10 times lower than that which was specified at the Wrangan training stage for FFHQ, and 7 times lower for LSUN Church. Despite this reduction, the proposed methodology is still successful.

\begin{figure*}[h!]
   \centering
    \includegraphics[width=1.0\linewidth]{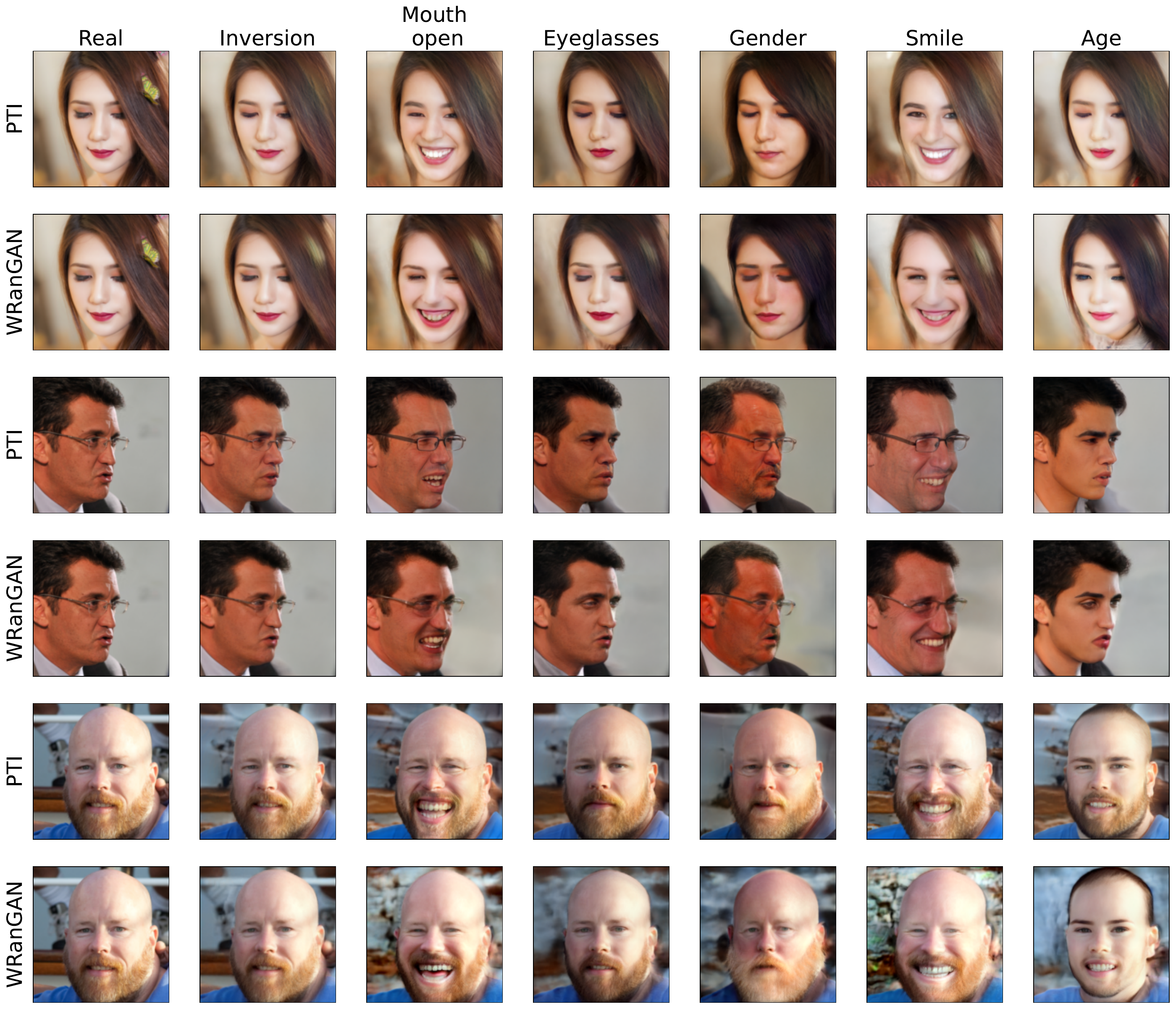}
    \caption{Qualitative editing comparisons for FFHQ dataset}
    \label{fig:add_editing_ffhq_1}
\end{figure*}

\begin{figure*}[h!]
   \centering
    \includegraphics[width=1.0\linewidth]{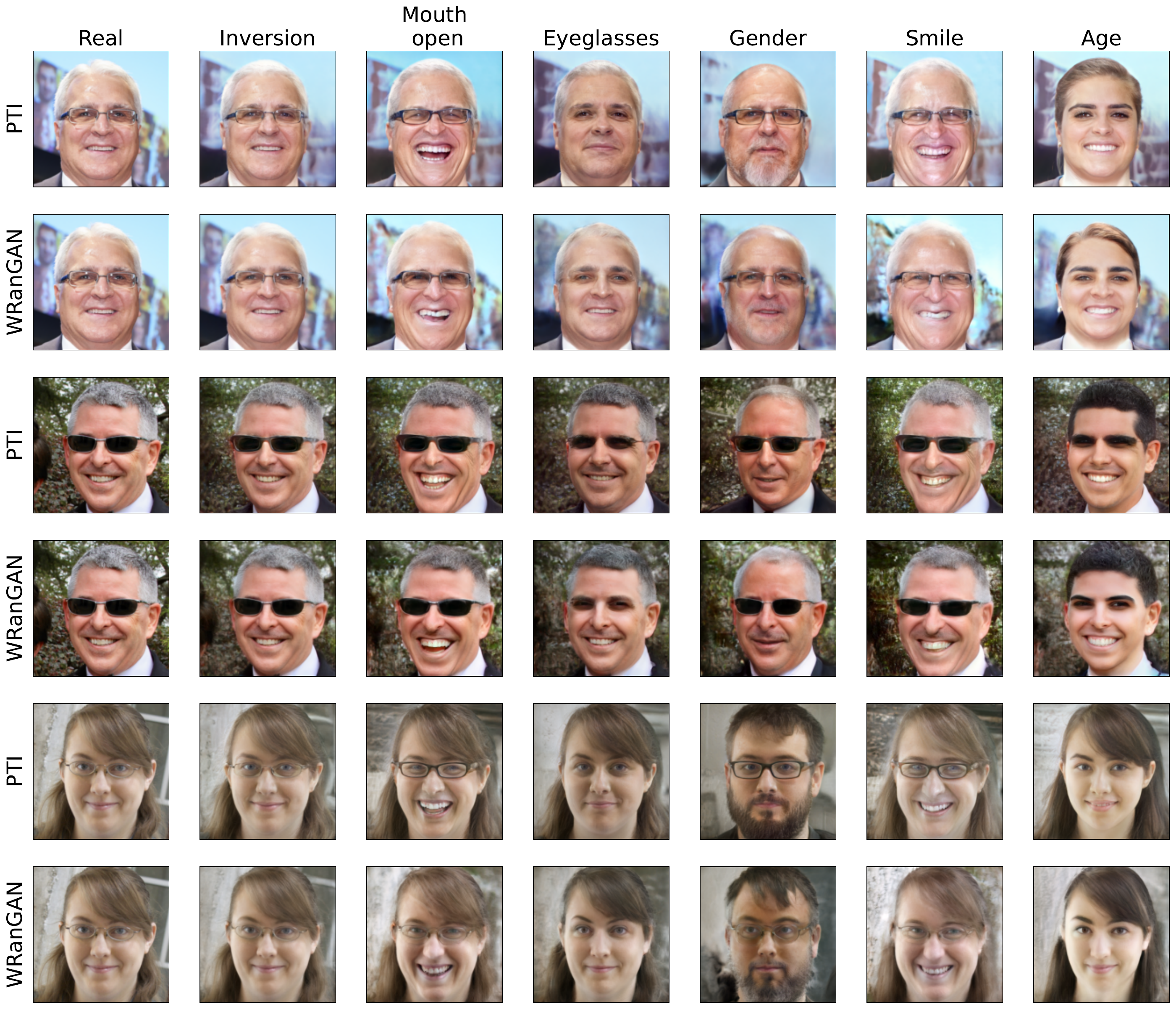}
    \caption{Qualitative editing comparisons for FFHQ dataset}
    \label{fig:add_editing_ffhq_2}
\end{figure*}

\begin{figure*}[h!]
   \centering
    \includegraphics[width=1.0\linewidth]{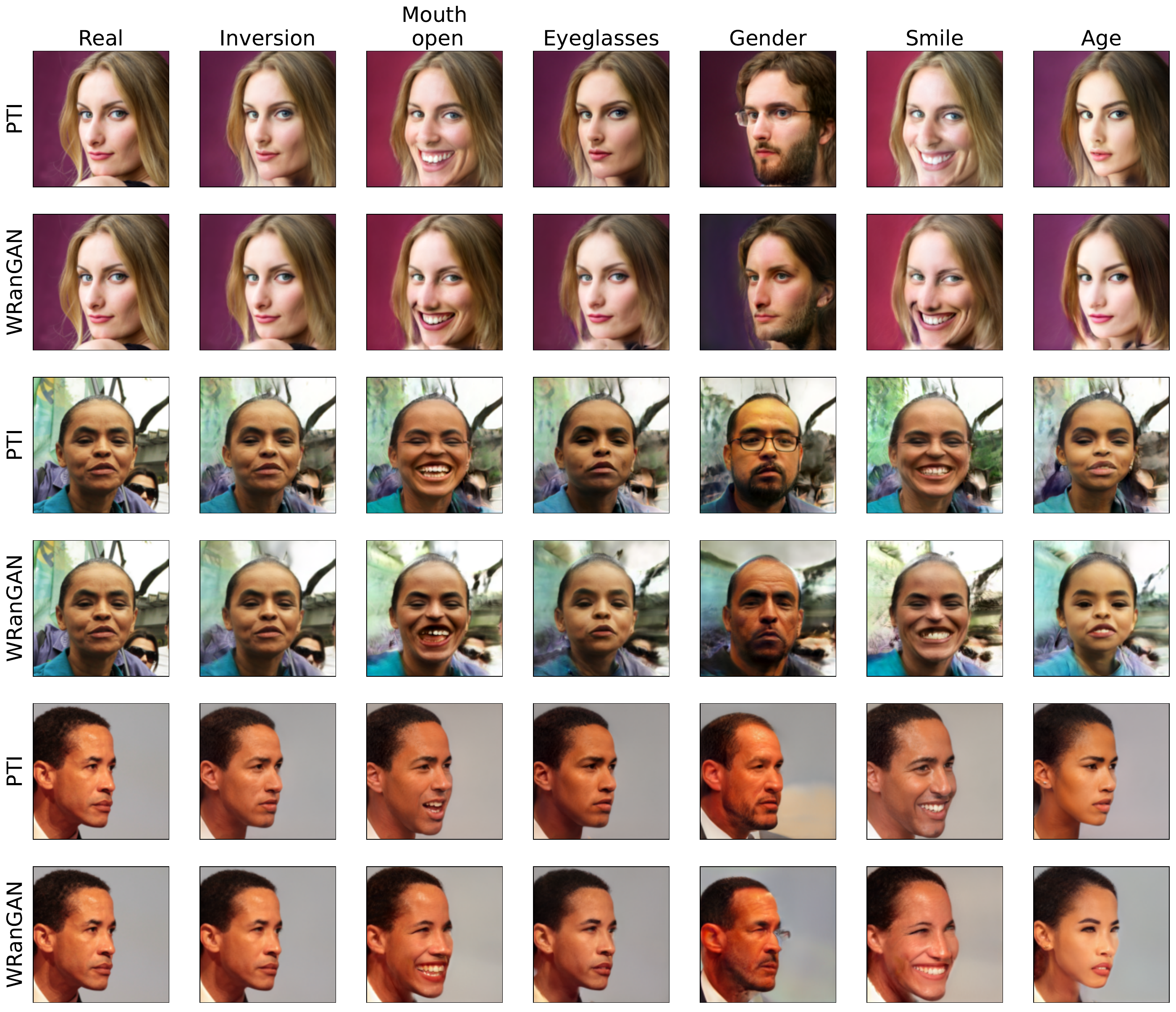}
    \caption{Qualitative editing comparisons for FFHQ dataset}
    \label{fig:add_editing_ffhq_3}
\end{figure*}

\begin{figure*}[h!]
   \centering
    \includegraphics[width=1.0\linewidth]{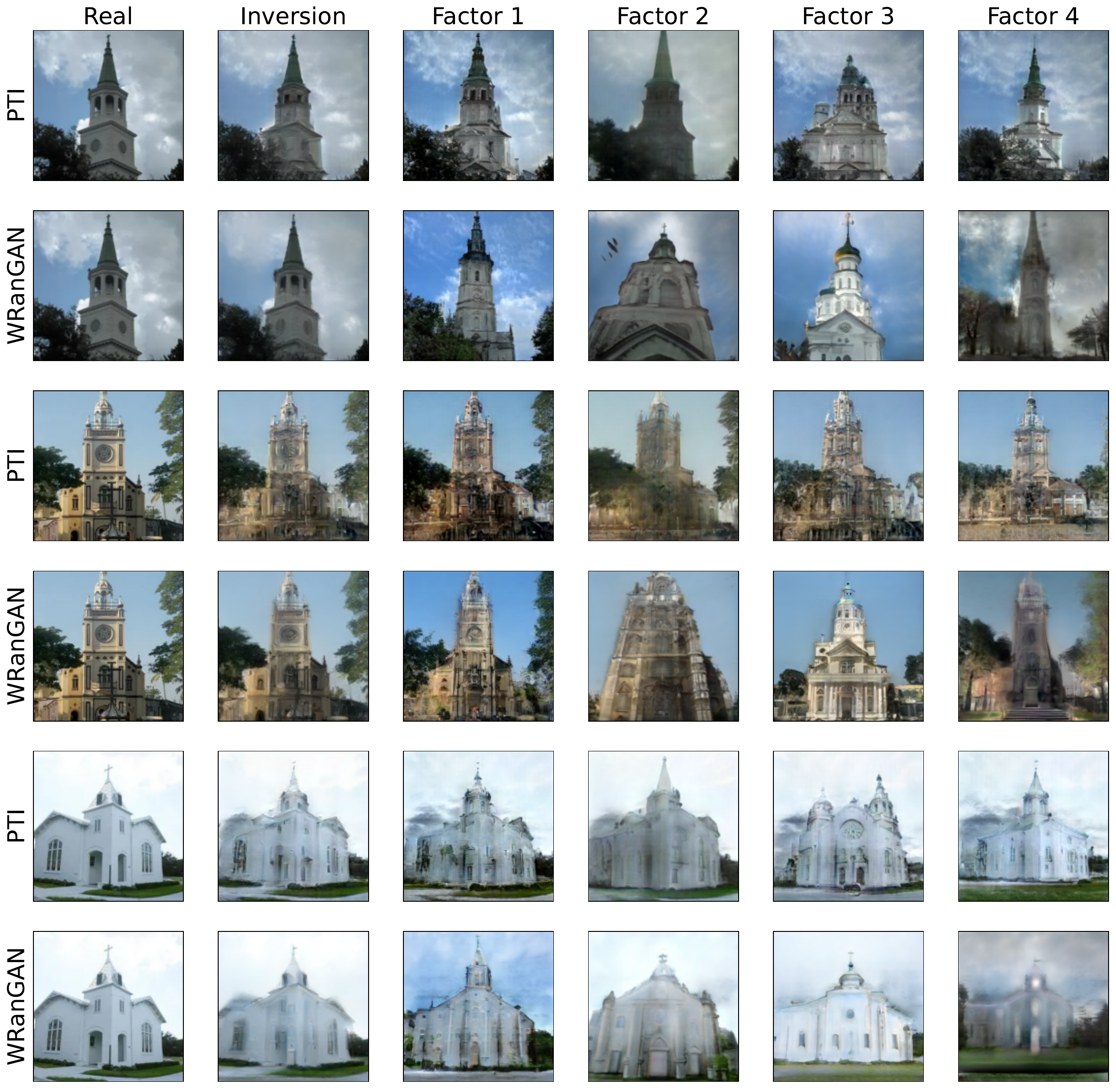}
    \caption{Qualitative editing comparisons for LSUN Church}
    \label{fig:add_editing_church_1}
\end{figure*}

\begin{figure*}[h!]
   \centering
    \includegraphics[width=1.0\linewidth]{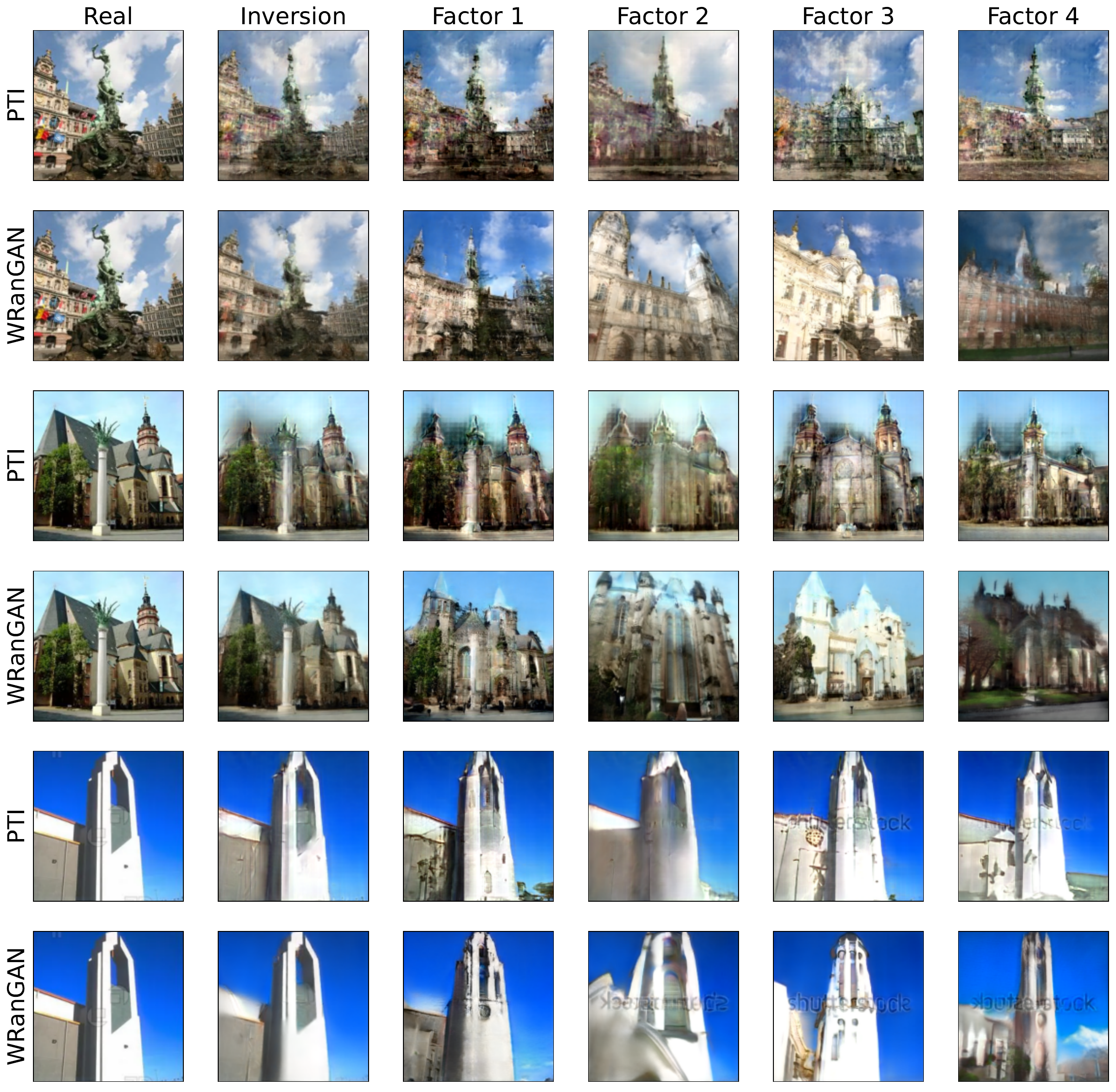}
    \caption{Qualitative editing comparisons for LSUN Church}
    \label{fig:add_editing_church_2}
\end{figure*}

\section{Additional qualitative comparisons for editing}

 We conducted additional comparative experiments of the proposed WRanGAN approach and the PTI inversion method for the StyleGAN 2 model in two domains: the FFHQ domain, with semantic directions corresponding to binary image attributes \cite{shen2020interpreting}, and the LSUN Church domain, with the first 4 vectors obtained by PCA approach \cite{disentangledrepr}. The results were captured in Figures ~\ref{fig:add_editing_ffhq_1},~\ref{fig:add_editing_ffhq_2},~\ref{fig:add_editing_ffhq_3},~\ref{fig:add_editing_church_1}, and~\ref{fig:add_editing_church_2}.

\end{document}